\documentclass[lettersize,journal]{IEEEtran}
\usepackage[T1]{fontenc}
\usepackage[hyphens]{url}
\usepackage{graphicx}
\usepackage{cite}
\usepackage{booktabs}
\usepackage{amsmath,amssymb}
\usepackage{multirow}
\usepackage[table]{xcolor}
\newif\ifreviewchanges
\reviewchangesfalse
\DeclareRobustCommand{\rev}[1]{{\ifreviewchanges\color{red}\fi#1}}
\newenvironment{revision}{\par\begingroup\clubpenalties 3 10000 10000 0\ifreviewchanges\color{red}\fi}{\par\endgroup}

\newif\ifcurrentreview
\currentreviewfalse
\DeclareRobustCommand{\newrev}[1]{{\ifcurrentreview\color{red}\fi#1}}
\newenvironment{newrevision}{\par\begingroup\clubpenalties 3 10000 10000 0\ifcurrentreview\color{red}\fi}{\par\endgroup}

\newif\iflayoutreview
\layoutreviewfalse
\DeclareRobustCommand{\fillrev}[1]{{\iflayoutreview\color{red}\fi#1}}

\newif\ifoctreview
\octreviewfalse
\DeclareRobustCommand{\octrev}[1]{{\ifoctreview\color{red}\fi#1}}

\newif\ifabstractcheck
\abstractcheckfalse
\DeclareRobustCommand{\absrev}[1]{{\ifabstractcheck\color{red}\fi#1}}

\newif\ifcamerareview
\camerareviewfalse
\DeclareRobustCommand{\camrev}[1]{{\ifcamerareview\color{red}\fi#1}}

\newif\ifdatasetreview
\datasetreviewfalse
\DeclareRobustCommand{\datasetrev}[1]{{\ifdatasetreview\color{blue}\hypersetup{linkcolor=blue,citecolor=blue}\fi#1}}

\newcommand{\rank}{Rank}

\renewcommand{\paragraph}[1]{\par\vspace{0.2em}\noindent\textbf{#1}\enspace\ignorespaces}

\usepackage{iftex}
\usepackage{combelow}
\ifPDFTeX
  \usepackage[utf8]{inputenc}
  \DeclareUnicodeCharacter{0219}{\cb{s}}
\else
  \catcode"0219=\active
  \protected\defș{\cb{s}}
\fi

\usepackage[labelsep=period]{caption}
\renewcommand{\paragraph}[1]{\par\vspace{0.2em}\noindent \textbf{#1 \hspace{0.2em}}\ignorespaces}

\usepackage[pass]{geometry}
\usepackage[breaklinks,colorlinks,citecolor=green,linkcolor=black,urlcolor=black]{hyperref}

\title{IntriAdapt:\,Generalizable\,Intrinsic-Guided\,Test-Time\\Adaptation for Metric Geometry Estimation}

\author{Jiaxing Chen,
        Zizheng Yan,~\camrev{\IEEEmembership{Member,~IEEE,}}
        Jingyu Zhuang,
        Jinwei Chen,\\
        Shuguang Cui,~\IEEEmembership{Fellow,~IEEE,}
        Guanying Chen,~\IEEEmembership{Member,~IEEE,}
        and Qingnan Fan}

\begin{document}
\addtocontents{toc}{\protect\setcounter{tocdepth}{-1}}

\IEEEaftertitletext{%
\begin{center}
  \vspace{-1em}
  \centering
\includegraphics[trim=1.6cm 9.55cm 1.6cm 10.7cm,clip,width=\linewidth]{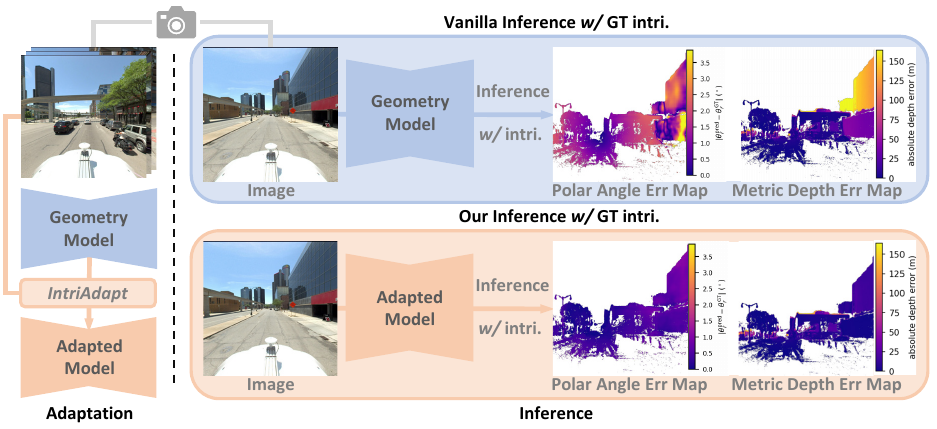}%
  \captionof{figure}{\rev{We propose \textit{IntriAdapt}, a test-time adaptation~(TTA) method for metric geometry estimation. \textit{Left}: Given a target image sequence, \textit{IntriAdapt} updates a LoRA adapter to align predicted point clouds with known camera intrinsics. \textit{Right}: The adapted weights generalize to novel images from the \emph{same} camera without per-frame optimization. As shown in the error maps, enforcing intrinsic consistency (polar-angle error) significantly reduces absolute metric depth error across all frames.}}
  \label{fig:teaser}
\end{center}
}
\maketitle
\raggedbottom
\noindent\begin{minipage}[t]{\columnwidth}
\setlength{\parindent}{1em}
\begin{abstract}
Recovering metric 3D geometry from images is fundamentally ill-posed due to perspective projection ambiguity. \absrev{Current geometry models incorporate supplied or predicted intrinsics during inference, but do not use them to supervise model adaptation. To refine these predictions, test-time adaptation (TTA) typically relies on external depth measurements. In this paper, we instead use camera intrinsics as an output-side geometric verifier. Comparing the point cloud's viewing rays with those defined by the known intrinsics yields a direct measure of geometric inconsistency. We propose \textit{IntriAdapt}, a TTA framework that minimizes this closed-form geometric residual.} By optimizing a lightweight LoRA adapter on a \absrev{frozen geometry foundation model}, our method aligns the prediction with the specific camera model without depth sensors or external priors. \absrev{We further extend this approach to video sequences, demonstrating that adapting a short sequence generalizes to other sequences captured by the same camera. The adapted model processes held-out images in a feed-forward pass, without repeating optimization for each frame. Experiments on ten single-image and seven sequence benchmarks across multiple base models show that \textit{IntriAdapt} consistently outperforms both unadapted and oracle-intrinsics baselines in metric accuracy and relative geometry.}

\end{abstract}

\end{minipage}\par
\section{Introduction}
Recovering metric 3D geometry from images alone with accurate real-world scale is important for various applications, from autonomous navigation to augmented reality~\cite{wang2025moge,wang2026moge2,lin2025da3,keetha2025mapanything}. The core difficulty is fundamental: perspective projection is ambiguous~\cite{wang2025moge,hu2024metric3dv2}. A scene at twice the distance, viewed through a camera with twice the focal length, produces a very similar image. Resolving this from visual features alone is ill-posed. \newrev{This matters when deploying a model across cameras: an image can support a visually plausible reconstruction while its predicted distances and viewing directions remain inconsistent with the acquisition setup. A useful metric estimator must therefore combine learned scene priors with the projection geometry of the target camera.}

\begin{figure*}[!t]
\input{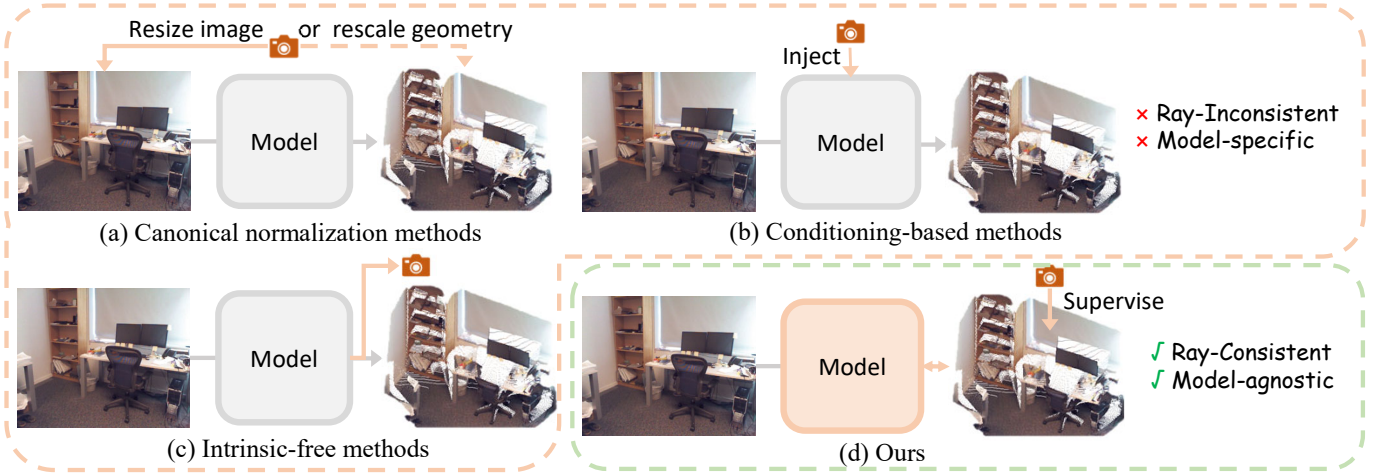}%
\caption{\camrev{Use of camera intrinsics. Canonical normalization methods~\cite{hu2024metric3dv2} rescale images or geometry; conditioning-based methods~\cite{keetha2025mapanything} inject intrinsics; intrinsic-free methods~\cite{wang2026moge2} infer geometry without supplied intrinsics. IntriAdapt uses known intrinsics to supervise adaptation.}}
\label{fig:method_pipes}
\end{figure*}

\newrev{Existing feed-forward methods incorporate camera information in several ways.} Fig.~\ref{fig:method_pipes} contrasts these uses with our output-side supervision.
\camrev{Canonical normalization methods}~\cite{yin2023metric3d,hu2024metric3dv2,lin2025da3} attribute metric ambiguity to focal length. They renormalize training data to a reference focal length $f^c$ by resizing images and rescaling depths, enabling generalization to unseen cameras.
Conditioning-based methods~\cite{piccinelli2024unidepth,piccinelli2025unidepthv2,keetha2025mapanything} go a step further by encoding intrinsics as explicit side information fed directly into the forward pass. \newrev{MapAnything~\cite{keetha2025mapanything}, for example, represents camera calibration as per-pixel viewing rays and combines this geometric input with image features.}
\newrev{\camrev{Intrinsic-free methods} instead infer camera or scale information from the image: Depth Pro~\cite{bochkovskiy2025depthpro} estimates focal length, while MoGe2~\cite{wang2026moge2} augments affine-invariant point prediction with a learned metric-scale estimate. These strategies make camera information useful for inference, but do not use the discrepancy between the predicted geometry and the known camera as a signal to adapt network weights.}

\newrev{Supplying accurate intrinsics and correcting the learned geometry are consequently distinct operations. A post-hoc camera solve can resolve global projection parameters while retaining errors in the underlying point map; similarly, intrinsic conditioning leaves the response to the supplied camera dependent on the pretrained model.} As shown in Fig.~\ref{fig:teaser}, although the geometry model (MoGe2) is trained on data covering a diverse range of camera intrinsics, and its oracle variant uses ground-truth intrinsics at inference, this still does not guarantee that the output point cloud is consistent with the ground-truth intrinsics, nor does it allow the model to fully exploit intrinsic information when estimating metric point map. The same figure also illustrates our adaptation setting: the left panel illustrates adapter training on the target-camera sequence, while the right panel shows its transfer to unseen images from the same camera without per-frame optimization. The paired polar-angle and metric-depth error maps show how improved intrinsic consistency accompanies better \octrev{metric depth estimates across the displayed examples}.

Test-time adaptation can substantially correct pretrained geometric priors at inference. \newrev{Marigold-DC~\cite{viola2025marigolddc} guides a pretrained diffusion model by jointly optimizing its depth latent and global scale and shift against sparse depth observations. CAPA~\cite{ke2026capa} freezes a 3D foundation model and updates LoRA or visual-prompt parameters using sparse geometric measurements. Both show that inference-time optimization can improve metric predictions, but their supervision requires external depth, obtained from sources such as LiDAR or image-based reconstruction. Such measurements are not always available when only RGB images and camera calibration are provided.} A structural gap remains: camera intrinsics encode the geometry of image formation, but these methods still rely on external depth measurements for adaptation. The known camera geometry itself remains underused as supervision for correcting metric predictions at test time.

Closing this gap requires reframing camera intrinsics as supervision. \textit{They act as a verifier}. Intrinsics define a geometric constraint that any predicted point cloud must satisfy: its induced per-pixel ray map should match the ray map of the known camera. Their disagreement gives a closed-form residual computable from image and camera alone, and minimizing this residual adapts the network end-to-end. \newrev{The target ray map is defined at every valid pixel, so the constraint does not require sparse depth samples, pixel correspondences, or a second view. It constrains viewing directions, while the pretrained model supplies the scene and metric priors.} Because modern metric geometry models couple ray direction and depth in learned geometric representations, correcting ray-map inconsistency reshapes the point cloud and improves metric geometry. Thus, intrinsics reused at the output become a self-supervisory signal for metric prediction.

To this end, we propose \textit{IntriAdapt}, which performs test-time adaptation guided by camera intrinsics. Given an input image from a known camera, a geometry estimation model predicts an affine-invariant point cloud and we can solve an implied ray map. Comparing it against the ray map derived from known intrinsics yields a residual that serves as self-supervised guidance at test time. We apply this signal at test time rather than during training, turning intrinsics from a training-time distribution into a stronger output-time verifier for the specific image. Rather than optimizing the output point cloud directly, we update model parameters so that the model's feature space serves as a geometric constraint and the adapted model can directly generalize at inference without repeated optimization. To avoid corrupting the pretrained priors, we freeze the backbone and train only a LoRA adapter to minimize \octrev{this intrinsic consistency residual on the target-camera images}.

Furthermore, we show that \textit{IntriAdapt} can be extended to video sequences without test-time training on every frame. \newrev{Frames captured with a fixed camera configuration share the same intrinsic constraints, allowing a single adapter to learn corrections that recur across images. We optimize this adapter on a small subset of frames and reuse it on held-out sequences from the same sensor. This separates the adaptation cost from subsequent feed-forward prediction and tests whether intrinsic supervision transfers beyond the images used for optimization.}

Our contributions are as follows:
\begin{itemize}
\item We identify a new role for camera intrinsics in metric geometry estimation: an output-side geometric verifier, complementary to their input-side roles in previous paradigms, where the use of intrinsics remains limited.
\item We propose \textit{IntriAdapt}, which leverages ray-map consistency as self-supervision for metric geometry estimation at test time, enabling depth-sensor-free TTA. We further extend \textit{IntriAdapt} to video sequences and demonstrate that training on only a small subset of frames is sufficient to generalize across the entire sequence.
\item On zero-shot metric benchmarks, our method consistently improves over baselines across \textit{10 single-image} and \textit{7 sequence} datasets, using no depth labels, and further generalizes across frames captured by the same camera. \newrev{Experiments with MoGe2, UniDepthV2, and MapAnything also demonstrate that the adaptation principle applies across different geometry representations.}
\end{itemize}

\section{Related Work}

\subsection{\newrev{Monocular Metric Depth Estimation}}
Recovering absolute depth from an image is a long-standing challenge. Early supervised methods struggled to transfer across environments~\cite{eigen2014depth,fu2018deep,yin2019vn,bhat2021adabins,guizilini2023towards}. Scaling to open-domain metric prediction has since become central. A persistent obstacle is that different datasets carry incompatible camera intrinsics, so naive multi-source training can learn inconsistent metric scale.
\begin{newrevision}
Relative-depth models address part of this problem by learning geometry without requiring a common absolute scale across training datasets. MiDaS~\cite{ranftl2020midas} develops mixed-dataset training for cross-dataset transfer, while Depth Anything and its successor~\cite{yang2024depth,yang2024depthv2} strengthen relative-depth priors through large-scale training. These priors support broad generalization, but their outputs require additional modeling or supervision to recover metric distances.

Metric estimators introduce different mechanisms to bridge this gap. ZoeDepth~\cite{bhat2023zoedepth} combines relative-depth pretraining with metric fine-tuning through lightweight metric-bin heads, allowing different domains to share a common representation. \camrev{Canonical normalization methods, including Metric3D and Metric3Dv2}~\cite{yin2023metric3d,hu2024metric3dv2}, normalize training samples into a canonical camera space to reduce focal-length-induced ambiguity. Depth Pro~\cite{bochkovskiy2025depthpro} combines a multi-scale vision transformer with focal-length estimation, producing sharp metric depth without requiring supplied camera metadata. These methods improve how image features and camera information determine metric depth in a forward pass.

Another approach supplies metric observations directly. Prompt Depth Anything~\cite{lin2025promptda} fuses low-cost LiDAR measurements into a depth decoder at multiple scales, while Prior Depth Anything~\cite{wang2025priorda} aligns and refines incomplete metric priors using a dense monocular prediction. These methods exploit complementary depth measurements to anchor scale and refine structure. IntriAdapt instead targets the setting where calibrated images are available without depth observations: it uses intrinsic consistency to adapt the learned geometric representation, and obtains metric depth from the resulting \octrev{adapted camera-space point map}.
\end{newrevision}

\subsection{\newrev{Point-Map and Camera Estimation}}
\begin{newrevision}
Estimating a full 3D point map requires recovering both scene depth and the camera geometry that lifts image pixels into space. LeReS~\cite{wei2021leres,yin2022accurate3dscene} combines depth estimation with camera-related recovery to reconstruct scene shape. \camrev{Conditioning-based methods, such as UniDepth and UniDepthV2}~\cite{piccinelli2024unidepth,piccinelli2025unidepthv2}, couple camera and depth estimation more closely: a dense camera representation conditions depth prediction, making the relation between viewing direction and scene distance part of the model. This formulation also provides a way to incorporate known intrinsics when available.

Point-map regression offers another route to reconstruction. DUSt3R~\cite{wang2024dust3r} predicts dense 3D correspondences through point maps, and subsequent models such as VGGT, Fast3R, and $\pi^3$~\cite{wang2025vggt,yang2025fast3r,wang2025pi3} extend learned reconstruction to multi-view inputs and camera estimation. For monocular images, MoGe~\cite{wang2025moge} predicts affine-invariant point maps and uses their projective structure to recover camera parameters and axial shift. MoGe2~\cite{wang2026moge2} adds metric-scale prediction while retaining this geometric representation. MapAnything~\cite{keetha2025mapanything} further combines images with optional geometric inputs, including camera rays, to recover metric multi-view geometry.

These models differ in how they represent camera information, but each exposes geometric quantities from which predicted viewing directions can be obtained. The common requirement is a predicted ray map that remains differentiably connected to the model parameters. IntriAdapt uses this connection to turn camera inconsistency into a weight-update signal, complementing the original models' feed-forward estimates and camera solvers.
\end{newrevision}

\begin{figure*}[!t]
\input{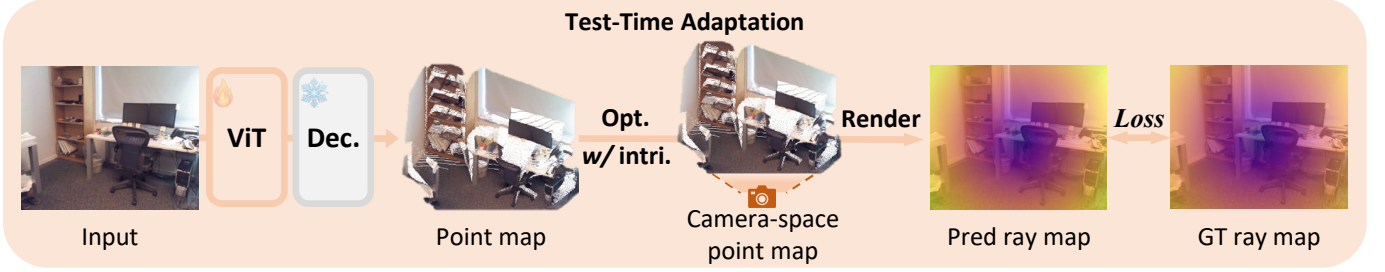}%
\caption{\rev{\textbf{Overview of \textit{IntriAdapt}.} Given an image and the ground-truth intrinsics, the frozen backbone (augmented with a LoRA adapter) predicts an affine invariant point map. A post-hoc optimization with ground-truth intrinsic recovers shift. The shift-compensated camera-space point cloud defines the predicted ray map, which is compared pixelwise against the ground-truth ray map.}}
\label{fig:method}
\end{figure*}

\subsection{Intrinsic-Guided Geometry Estimation}
Camera intrinsics are fundamental in classical multiple-view geometry, where reprojection consistency jointly optimizes cameras, structure, and poses~\cite{hartley2004multiple,triggs2000bundle}. These works establish projection geometry as a powerful constraint, but their optimized variables are geometric quantities rather than the parameters of a point-map prediction network. \newrev{The resulting correction is represented in a particular reconstruction; it does not by itself produce an adapted predictor for new images.} Learning-based self-supervised depth also uses intrinsics, but mainly as projection operators for multi-view photometric reconstruction; supervision comes from RGB consistency rather than intrinsics themselves~\cite{zhou2017sfmlearner,godard2019monodepth2}.
Unlike these methods, our formulation uses known camera intrinsics as the supervision target. \newrev{A calibrated pixel grid defines a dense ray map independently of scene appearance, camera motion, and predicted depth. Comparing it with the model's rays therefore provides supervision even for a single image \octrev{captured with known camera intrinsics}.}

\par\csname @IEEEtranneedspace\endcsname{4\baselineskip}{\relax}
\subsection{Test-time Adaptation}
\begin{newrevision}
Test-time adaptation uses observations available at deployment to refine a pretrained predictor. In depth completion, Park et al.~\cite{park2024test} use sparse-depth features as proxies for source-domain representations. ETA~\cite{chung2025eta} instead learns an energy function on source data and uses it to guide updates of a completion network on shifted test inputs. In the related continual-learning setting, ProtoDepth~\cite{Rim_2025_CVPR} learns domain-specific prototypes that adapt the latent features of a frozen model while reducing forgetting. These approaches address domain shift in systems that receive sparse depth alongside \octrev{the input RGB images at inference time}.

Optimization can also act on a generative model's latent variables or on the alignment of a predicted depth map. Marigold-DC~\cite{viola2025marigolddc} steers diffusion using sparse depth residuals, jointly adjusting the depth latent and global scale and shift during denoising. SteeredMarigold~\cite{gregorek2025steeredmarigold} conditions diffusion on available depth points to complete largely incomplete maps, while zero-shot test-time alignment~\cite{hyoseok2025zero} adapts an affine-invariant depth prior to metric observations. These methods demonstrate the value of pretrained priors, with measured depths supplying the missing metric constraints.

CAPA~\cite{ke2026capa} is close to our parameter-efficient setting. It freezes a 3D foundation model and optimizes LoRA or visual prompts using sparse geometric cues; for videos, it shares adapted parameters across frames. IntriAdapt changes the source of supervision. Its residual compares predicted rays with those determined by known intrinsics, without requiring measured depth. Furthermore, the learned adapter can transfer to held-out
frames or sequences from the same sensor.
\end{newrevision}

\section{Method}

\rev{We now formalize \textit{IntriAdapt}, a test-time adaptation framework in which intrinsics play the role of an \emph{output-side geometric verifier}. We present a model-agnostic affine-invariant point map view: a dense point cloud produced by any geometry estimator that is not strictly calibrated can be treated as a generalized affine-invariant prediction. Thus, any geometry estimation model can be corrected by our framework. An overview of \textit{IntriAdapt} is shown in Fig.~\ref{fig:method}.}

\subsection{Preliminaries}
\label{sec:prelim}

We first review affine-invariant point-map and the post-hoc optimization used to recover intrinsics and axial shift.

\paragraph{Affine-invariant point map.}
Given an RGB image $\mathbf{I}\in\mathbb{R}^{H\times W\times 3}$, a geometry model $f_\theta$ predicts a dense point map
\begin{equation}
    \hat{\mathbf{P}} = f_\theta(\mathbf{I})\in\mathbb{R}^{H\times W\times 3},
\end{equation}
expressed in the camera's coordinate frame. Following MoGe~\cite{wang2025moge,wang2026moge2}, the supervision is affine-invariant: at training time the network is only required to recover $\hat{\mathbf{P}}$ up to a global scale and translation with respect to the ground-truth cloud $\mathbf{P}$. A metric reconstruction $\tilde{\mathbf{P}}=s\,(\hat{\mathbf{P}}+\mathbf{t})$ therefore requires a scale $s\in\mathbb{R}^{+}$ and a shift $\mathbf{t}\in\mathbb{R}^3$ to be recovered at inference. 

\begin{figure}[!t]
\input{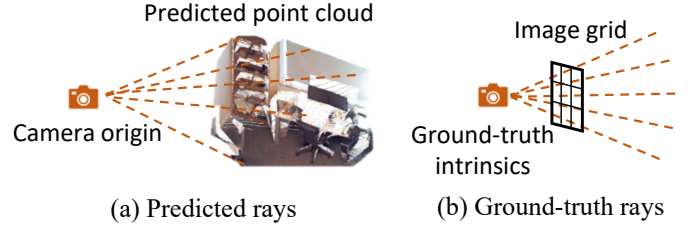}%
\caption{\camrev{Predicted rays from the shift-compensated camera origin and predicted point cloud (left); ground-truth rays from intrinsics (right).}}
\label{fig:gen_ray}
\end{figure}

\paragraph{Post-hoc intrinsic and shift optimization.}
Under a centered principal point and square-pixel assumption, the shift can be reduced to a single axial component $\mathbf{t}=(0,0,\tau)$ without loss of generality. Given the image grid $\{(u_i,v_i)\}$, MoGe recovers $(\tau, f)$ jointly by enforcing a projection constraint between the shifted point cloud and its 2D locations:
\begin{equation}
    (\tau^{*}, f^{*}) = \arg\min_{\tau,f}\; \sum_{i\in\mathcal{M}} \rho\!\left( \pi_{f}\big(\hat{\mathbf{p}}_i + (0,0,\tau)\big) - (u_i, v_i) \right),
    \label{eq:moge_opt}
\end{equation}
where $\pi_{f}(\cdot)$ denotes perspective projection with focal length $f$, $\rho(\cdot)$ is a robust reprojection loss, and $\mathcal{M}$ is the set of valid pixels. This solver treats intrinsics as an \emph{output} of the reconstruction pipeline, computed once from the frozen prediction. Any systematic bias already baked into $\hat{\mathbf{P}}$ is absorbed into $f^{*}$ rather than being corrected. Our goal is to reuse intrinsics in the opposite direction: when $f$ is known, this very optimization becomes an oracle to audit $\hat{\mathbf{P}}$. With intrinsics fixed, the remaining discrepancy measures camera consistency across the image and provides the supervision for updating the geometry predictor.

\subsection{Ray-Map Residual from Known Intrinsics}
\label{sec:raymap}

Fig.~\ref{fig:gen_ray} illustrates the predicted and ground-truth rays used to form the intrinsic consistency residual. \fillrev{Their shared pixel grid enables per-pixel comparison.}

\paragraph{Ground-truth ray map.}
For a camera with ground-truth intrinsic matrix $\mathbf{K}_\text{gt}$ and image resolution $(H,W)$, every pixel $(u,v)$ defines a unique unit-norm viewing direction in the camera frame~\cite{zhang2024raydiffusion},
\begin{equation}
    \mathcal{R}_\text{gt}(u,v) \;=\; \frac{\mathbf{K}_\text{gt}^{-1}\,[u,v,1]^\top}{\|\mathbf{K}_\text{gt}^{-1}\,[u,v,1]^\top\|}\,.
    \label{eq:ray_gt}
\end{equation}
We call $\mathcal{R}_\text{gt}\in\mathbb{S}^{2}\subset\mathbb{R}^{H\times W\times 3}$ the \emph{ground-truth ray map}. It is a closed-form function of the intrinsics and the pixel grid alone, and carries no dependence on the scene or the network. 
\paragraph{Predicted ray map.}
Any dense prediction $\hat{\mathbf{P}}$ also defines a ray map, once the axial shift $\tau$ is resolved. Concretely, after compensating for shift the point from the camera origin to pixel $(u,v)$ becomes $\hat{\mathbf{p}}_{u,v}+(0,0,\tau)$, and the corresponding ray direction is
\begin{equation}
    \mathcal{R}_\text{pred}(u,v;\,\tau,\theta) \;=\; \frac{\hat{\mathbf{p}}_{u,v}+(0,0,\tau)}{\|\hat{\mathbf{p}}_{u,v}+(0,0,\tau)\|}\,.
    \label{eq:ray_pred}
\end{equation}
In contrast to Eq.~\eqref{eq:ray_gt}, $\mathcal{R}_\text{pred}$ depends on the network parameters $\theta$ through $\hat{\mathbf{P}}$ and on the resolved shift $\tau$ through Eq.~\eqref{eq:moge_opt}. It therefore carries the \emph{shape} information of the prediction.

\paragraph{Intrinsics as a verifier.}
If the prediction were perfectly consistent with the camera that produced the image, the two ray maps would coincide everywhere. Any disagreement is by construction a geometric defect of $\hat{\mathbf{P}}$. We quantify this discrepancy with a pixel-wise cosine residual
\begin{equation}
    \mathcal{L}_{\text{ray}}(\theta) \;=\; \frac{1}{|\mathcal{M}|}\sum_{(u,v)\in\mathcal{M}}\Big[1\;-\;\mathcal{R}_\text{pred}(u,v;\tau^{*},\theta)\cdot\mathcal{R}_\text{gt}(u,v)\Big],
    \label{eq:ray_loss}
\end{equation}
where $\tau^{*}$ is obtained by solving Eq.~\eqref{eq:moge_opt} with the known $f=f_\text{gt}$ held fixed. Because $\mathcal{R}_\text{gt}$ is independent of the scene, $\mathcal{L}_{\text{ray}}$ can be computed from image and intrinsics alone, with no auxiliary prior. In this sense camera intrinsics act as a self-supervisory \emph{verifier} defined at the output of the network, dual to the input-side role they play in previous methods.
\paragraph{Gradient flow.}
Eq.~\eqref{eq:ray_loss} is differentiable in the network parameters $\theta$ through $\hat{\mathbf{P}}$. The axial shift $\tau^{*}$ enters as a solution of an optimization that we detach from the computation graph.
Although $\mathcal{L}_{\text{ray}}$ is defined on ray directions, its gradients update the full point map. In point-map models, ray direction and depth are encoded by the same learned geometric representation. Reducing ray inconsistency therefore reshapes $\hat{\mathbf{P}}$ and improves metric depth as well as point geometry.
\paragraph{Output-time reprojection.}
A second, simpler way to use $\mathbf{K}_\text{gt}$ at the \emph{output} stage, available to any depth model, is \emph{reprojection}. Once the resolved depth $\hat{\mathbf{Z}}(u,v)$ has been fixed (after any weight update and any post-hoc solve), \rev{each predicted 3D point is replaced with the on-ray point at the same depth:
\begin{equation}
\hat{\mathbf{p}}_{u,v}^{\mathrm{re}} = \hat{\mathbf{Z}}(u,v)\,
\frac{\mathcal{R}_\mathrm{gt}(u,v)}{[\mathcal{R}_\mathrm{gt}(u,v)]_z}.
\label{eq:depth_preserving_reprojection}
\end{equation}
The division by the ray's $z$ component preserves the depth coordinate because $\mathcal{R}_\mathrm{gt}$ is a unit-length direction.}
The resulting cloud is, by construction, perfectly consistent with the camera intrinsics. Although reprojection does provide a free improvement on top of an oracle-FoV prediction, it operates on a strictly different axis from \textit{IntriAdapt}: reprojection only redistributes points along their correct rays at their predicted depth, so it cannot correct biased depth values themselves and leaves the structural error of $\hat{\mathbf{P}}$ intact. \textit{IntriAdapt}, in contrast, modifies the depth values produced by the network, and the two operations therefore compose. Consequently, our final inference pipeline first performs test-time adaptation to improve the predicted geometry, and then applies output-time reprojection as a final geometry-preserving refinement.

\subsection{LoRA-based Test-Time Adaptation}
\label{sec:lora}

Naively updating all parameters of $f_\theta$ to minimize
$\mathcal{L}_{\text{ray}}$ risks damaging the pretrained geometric prior.
We therefore freeze the backbone and attach a low-rank adapter
(LoRA~\cite{hu2022lora}) to the encoder's attention layers, writing
$\theta=\theta_0+\Delta\theta_{\text{LoRA}}$, where
$\Delta\theta_{\text{LoRA}}$ are the only trainable variables.
We insert LoRA into the encoder rather than the prediction head. As a result, the same intrinsic consistency objective can be applied more naturally to geometry foundation
models with different prediction heads or output parameterizations.

At deployment, given a test image (or images sharing the same intrinsics), we solve
\begin{equation}
    \Delta\theta_{\text{LoRA}}^{*} \;=\; \arg\min_{\Delta\theta_{\text{LoRA}}}\; \mathcal{L}_{\text{ray}}\big(\theta_0+\Delta\theta_{\text{LoRA}}\big)
    \label{eq:tta}
\end{equation}
with a few steps. No ground-truth depth is ever used, and the only supervisory signal is the intrinsic-induced residual.

\subsection{Single-Image and Sequence Adaptation}
\label{sec:granularity}
\paragraph{Single-image adaptation.}
For an isolated test image, we perform $K{=}10$ optimization steps of Eq.~\eqref{eq:tta}, then generate the adapted prediction with $\theta_0+\Delta\theta_{\text{LoRA}}^{*}$ in a forward pass.
\paragraph{Sequence-level adaptation.}
When a video or an image set is captured by a single physical sensor, all frames share the same $\mathbf{K}_\text{gt}$, and the systematic component of the prediction error is therefore predominantly a camera-level bias rather than a per-frame correction. We exploit this by \emph{sharing} a single LoRA adapter across all frames of that sensor given $k$ adaptation sequences. 
We refer to this regime as \emph{Seq-$k$} adaptation. After the adapter is trained, the same $\Delta\theta_{\text{LoRA}}^{*}$ is then applied without any further update to all evaluation sequences captured by the same sensor. Our sequence experiments show that even Seq-$1$, a single short adaptation sequence, transfers across the rest of the benchmark and almost saturates the gain achievable from sensor-level supervision, casting ray-map supervision not as an instance-level fix but as an intrinsic-specific fine-tuning signal that pays off once the sensor is known.
\paragraph{Oracle variant.}
Throughout the experiments, a superscript $^{*}$ denotes access to ground-truth intrinsics at test time. For non-adapted baselines, there are two architecture-dependent ways to consume the intrinsics. For intrinsic-conditioned models such as UniDepthV2 and MapAnything, the ground-truth intrinsics are fed directly to the forward pass as side information. For other point-map models such as MoGe2, the ground-truth focal length is used in the post-hoc solve of Eq.~\eqref{eq:moge_opt}. Ours$^{*}$ uses the same intrinsics as a self-supervised adaptation signal, and can improve the shape of $\hat{\mathbf{P}}$. The resulting depth from all these methods can be reprojected onto the ground-truth rays as described above.

\section{Experiments}
\label{sec:exp}

We evaluate \textit{IntriAdapt} under single-image adaptation, sequence-level adaptation, and targeted ablations.

\subsection{Setup}
\label{sec:exp_setup}

\begin{table}[!t]
\centering
\small
\setlength{\tabcolsep}{2pt}
\caption{\rev{Sequence benchmark statistics after intrinsic grouping.}}
\label{tab:seq_summary_main}
\begin{tabular*}{\linewidth}{@{\extracolsep{\fill}}lrrrrr@{}}
\toprule
Dataset & \shortstack{FoVx\\($^\circ$)} & \shortstack{FoVy\\($^\circ$)} & \#Seqs & \shortstack{Frames\\/Seq} & \#Frames \\
\midrule
DIODE       & 42.3 & 32.4 & 20 & 38.5  & 771 \\
KITTI\_09\_26 & 54.3 & 28.8 & 21 & 23.2 & 487 \\
7-Scenes    & 57.4 & 44.6 & 18 & 100.0 & 1800 \\
ScanNet     & 57.9 & 45.0 & 56 & 100.0 & 5600 \\
HAMMER      & 63.7 & 34.5 & 29 & 26.7  & 775 \\
ETH3D       & 84.3 & 62.2 & 12 & 35.7  & 428 \\
Metropolis  & 90.0 & 90.0 & 35 & 95.2  & 3332 \\
\bottomrule
\end{tabular*}

\end{table}

\paragraph{Benchmarks.}
For single-image evaluation, we follow the MoGe2~\cite{wang2026moge2} protocol and report averages over the same $10$ zero-shot benchmarks. For sequence evaluation, we use seven camera-indexed benchmarks: three video datasets from the CAPA~\cite{ke2026capa} split (7-Scenes, ScanNet, Metropolis) and four high-diversity image benchmarks (ETH3D, KITTI, DIODE, HAMMER) re-organized by physical sensor. Since ScanNet overlaps with the training sets of some compared baselines, the main paper reports the leakage-free $6$-dataset average, with the full $7$-dataset result deferred to the supplementary material. \rev{Table~\ref{tab:seq_summary_main} summarizes the constructed sequence benchmark: $191$ sequences and $13{,}193$ frames in total, or $135$ sequences and $7{,}593$ frames when ScanNet is excluded. These are benchmark-composition counts before selecting adaptation and held-out sequences. The processed horizontal FoV spans $42.3^\circ$--$90.0^\circ$, and the average sequence length ranges from $23.2$ to $100$ frames. This combines dense video coverage with shorter, more diverse image collections. Images are grouped using a tolerance of $0.005$ on normalized intrinsic entries; KITTI denotes the KITTI\_09\_26 camera group. The supplementary material retains the full table, including scene type and resolution, together with the grouping, cleaning, and \octrev{sampling details for every sequence benchmark}.}

\begin{revision}
\paragraph{Implementation.}
The main experiments use MoGe2~\cite{wang2026moge2} as the frozen base model. We insert rank-$16$ LoRA adapters into every attention layer of its DINOv2 encoder, giving $1.57$M trainable parameters, approximately $0.4\%$ of the backbone weights. We optimize the pixelwise ray-map residual with AdamW at a learning rate of $3{\times}10^{-5}$. Single-image adaptation resets the adapter for every image and runs $10$ optimization steps \octrev{before producing the final adapted prediction}.

For Seq-$k$ adaptation, $k$ denotes the number of sequences captured by the target camera used to train one shared adapter. We then evaluate it on held-out sequences from the same sensor without further updates. The default is Seq-$1$ with $20$ AdamW updates. At each update, a mini-batch covers $50\%$ of the available adaptation frames; gradients are accumulated over four sub-batches before one optimizer step. Ray maps are evaluated on a $64{\times}64$ grid. Metrics are computed per image and then averaged, without temporal aggregation. All sequence methods use the same held-out sequences within each comparison \octrev{to keep the evaluation split fixed}.

We also adapt UniDepthV2~\cite{piccinelli2025unidepthv2} and MapAnything~\cite{keetha2025mapanything}, using the same LoRA rank, optimizer, learning rate, adaptation steps, ray-map resolution, and held-out protocol. UniDepthV2 follows the MoGe2 batching strategy. For the multi-view MapAnything model, each optimizer step instead uses one batch of four views, without accumulating four sub-batches. Other main results use MoGe2 unless stated otherwise.
\end{revision}

\begin{table*}[!t]
\centering
\small
\setlength{\tabcolsep}{2.5pt}
\caption{\rev{Single-image geometry estimation averaged over ten benchmarks.}}
\label{tab:sf_main}
\resizebox{\textwidth}{!}{%
\begin{tabular}{l|ccc|ccc|ccc|ccc|ccc|ccc|c}
\toprule
\multirow{3}{*}{Method} & \multicolumn{9}{c|}{\textit{Point}} & \multicolumn{9}{c|}{\textit{Depth}} & Avg. \\
\cmidrule(lr){2-10}\cmidrule(lr){11-19}
 & \multicolumn{3}{c|}{Scale-inv.} & \multicolumn{3}{c|}{Affine-inv.} & \multicolumn{3}{c|}{Metric} & \multicolumn{3}{c|}{Scale-inv.} & \multicolumn{3}{c|}{Affine-inv.} & \multicolumn{3}{c|}{Metric} & \rank$\downarrow$ \\
 & Rel$\downarrow$ & $\delta^{1.03}\!\uparrow$ & \rank$\downarrow$ & Rel$\downarrow$ & $\delta^{1.03}\!\uparrow$ & \rank$\downarrow$ & Rel$\downarrow$ & $\delta^{1.03}\!\uparrow$ & \rank$\downarrow$ & Rel$\downarrow$ & $\delta^{1.03}\!\uparrow$ & \rank$\downarrow$ & Rel$\downarrow$ & $\delta^{1.03}\!\uparrow$ & \rank$\downarrow$ & Rel$\downarrow$ & $\delta^{1.03}\!\uparrow$ & \rank$\downarrow$ & \\
\midrule
DA3 & - & - & - & - & - & - & - & - & - & 9.25 & 45.71 & 6.85 & 7.44 & 55.86 & 6.85 & \underline{12.78} & 22.08 & \underline{2.79} & 5.50 \\
MapAnything & 10.95 & 25.36 & 6.00 & 8.31 & 45.03 & 6.20 & 16.34 & 14.24 & 6.57 & 8.08 & 51.34 & 6.30 & 6.55 & 59.78 & 7.00 & 27.43 & 6.96 & 7.36 & 6.57 \\
UniDepthV2 & 11.66 & 24.09 & 6.10 & 8.57 & 46.91 & 4.90 & 10.09 & 28.67 & 4.36 & 8.63 & 53.20 & 4.25 & 6.43 & 63.88 & 4.20 & 21.28 & 13.53 & 6.21 & 5.00 \\
MoGe2 & 10.73 & 28.47 & 5.00 & 7.85 & 49.57 & 3.80 & 7.95 & 34.39 & 3.29 & \textbf{7.30} & \underline{55.75} & \textbf{2.60} & \textbf{5.58} & \textbf{64.52} & \textbf{2.50} & 15.33 & 14.88 & 4.29 & 3.58 \\
\midrule
MapAnything$^{*}$ & \textbf{8.12} & 48.77 & 3.70 & \textbf{7.30} & 53.11 & 4.65 & 14.68 & 17.19 & 6.00 & \underline{7.75} & 53.88 & 5.40 & 6.38 & 60.59 & 5.95 & 27.46 & 8.81 & 6.57 & 5.38 \\
UniDepthV2$^{*}$ & \underline{8.55} & \underline{53.27} & \underline{2.45} & \underline{7.52} & \underline{54.55} & \underline{2.65} & 9.35 & \underline{35.35} & 3.29 & 8.37 & 54.74 & \underline{3.05} & 6.40 & \underline{64.21} & 3.75 & 18.48 & 21.38 & 3.86 & 3.17 \\
MoGe2$^{*}$ & 9.10 & 51.86 & 2.90 & 8.02 & 50.37 & 3.20 & \underline{7.94} & 34.95 & \underline{3.00} & 9.08 & 51.99 & 4.40 & \textbf{5.58} & \textbf{64.52} & \underline{2.60} & 13.14 & \underline{22.78} & 2.86 & \underline{3.16} \\
\midrule
Ours$^{*}$ & 8.56 & \textbf{56.02} & \textbf{1.85} & 7.73 & \textbf{54.96} & \textbf{2.60} & \textbf{7.26} & \textbf{39.99} & \textbf{1.50} & 8.55 & \textbf{56.06} & 3.15 & \underline{5.87} & 64.13 & 3.15 & \textbf{12.11} & \textbf{25.02} & \textbf{2.07} & \textbf{2.39} \\
\bottomrule
\end{tabular}
    }

\end{table*}

\begin{table*}[!t]
\centering
\small
\setlength{\tabcolsep}{2.5pt}
\caption{\rev{Sequence geometry estimation averaged over six benchmarks, excluding ScanNet.}}
\label{tab:seq6_main}
\resizebox{\textwidth}{!}{%
\begin{tabular}{l|ccc|ccc|ccc|ccc|ccc|ccc|c}
\toprule
\multirow{3}{*}{Method} & \multicolumn{9}{c|}{\textit{Point}} & \multicolumn{9}{c|}{\textit{Depth}} & Avg. \\
\cmidrule(lr){2-10}\cmidrule(lr){11-19}
 & \multicolumn{3}{c|}{Scale-inv.} & \multicolumn{3}{c|}{Affine-inv.} & \multicolumn{3}{c|}{Metric} & \multicolumn{3}{c|}{Scale-inv.} & \multicolumn{3}{c|}{Affine-inv.} & \multicolumn{3}{c|}{Metric} & \rank $\downarrow$ \\
 & Rel$\downarrow$ & $\delta^{1.03}\!\uparrow$ & \rank $\downarrow$ & Rel$\downarrow$ & $\delta^{1.03}\!\uparrow$ & \rank $\downarrow$ & Rel$\downarrow$ & $\delta^{1.03}\!\uparrow$ & \rank $\downarrow$ & Rel$\downarrow$ & $\delta^{1.03}\!\uparrow$ & \rank $\downarrow$ & Rel$\downarrow$ & $\delta^{1.03}\!\uparrow$ & \rank $\downarrow$ & Rel$\downarrow$ & $\delta^{1.03}\!\uparrow$ & \rank $\downarrow$ & \\
\midrule
VideoDA & - & - & - & - & - & - & - & - & - & - & 20.05 & 10.83 & - & 35.49 & 10.83 & - & - & - & 10.83 \\
\midrule
DA3 & - & - & - & - & - & - & - & - & - & 8.53 & 47.02 & 7.92 & 7.68 & 54.95 & 8.00 & 15.94 & 17.67 & 3.83 & 6.58 \\
UniDepthV2 & 12.42 & 27.30 & 6.25 & 10.45 & 45.97 & 5.42 & 15.83 & 22.03 & 5.33 & 8.69 & 55.72 & 4.67 & 7.92 & 61.35 & 4.83 & 29.74 & 11.56 & 6.50 & 5.50 \\
MoGe2 & 10.09 & 15.35 & 6.75 & 7.20 & 45.22 & 5.67 & 9.48 & 27.66 & 4.25 & \underline{6.53} & 54.84 & \underline{4.17} & \underline{5.56} & 63.16 & \underline{3.92} & 16.42 & 11.54 & 5.50 & 5.04 \\
\midrule
VGGT & 19.46 & 11.39 & 7.58 & 11.75 & 21.58 & 7.83 & - & - & - & 7.80 & 45.75 & 7.58 & 6.84 & 50.31 & 8.33 & - & - & - & 7.83 \\
MapAnything & 12.11 & 16.85 & 7.00 & 8.89 & 39.62 & 6.83 & 60.63 & 10.16 & 7.00 & 8.88 & 49.60 & 6.75 & 6.88 & 56.84 & 7.00 & 75.13 & 5.80 & 7.75 & 7.06 \\
DA3-Nested & 10.87 & 41.80 & 4.75 & 9.71 & 48.51 & 4.67 & 12.31 & 27.36 & 3.83 & 8.06 & \textbf{58.73} & 4.50 & 7.17 & \textbf{63.56} & 4.83 & 21.53 & 12.71 & 4.50 & 4.51 \\
\midrule
UniDepthV2$^{*}$ & 8.59 & \textbf{54.49} & \underline{2.17} & 8.18 & \textbf{55.99} & \textbf{2.67} & 12.66 & 29.15 & 4.42 & 8.29 & \underline{56.35} & \textbf{3.50} & 7.66 & 61.29 & 4.25 & 22.14 & 18.38 & 4.17 & \underline{3.53} \\
MoGe2$^{*}$ & \underline{7.32} & 49.08 & 3.58 & \underline{6.79} & 49.22 & 4.58 & \underline{9.25} & \underline{30.71} & \underline{3.42} & 7.30 & 49.19 & 6.33 & \underline{5.56} & \underline{63.17} & \textbf{3.58} & \underline{13.92} & \underline{19.57} & \underline{3.58} & 4.18 \\
MapAnything$^{*}$ & 9.24 & 40.75 & 4.83 & 7.79 & 50.69 & 4.50 & 48.41 & 13.74 & 6.00 & 8.63 & 51.53 & 5.42 & 6.82 & 57.41 & 6.25 & 67.88 & 7.28 & 7.42 & 5.74 \\
\midrule
Ours$^{*}$ & \textbf{6.43} & \underline{54.45} & \textbf{2.08} & \textbf{6.12} & \underline{55.05} & \underline{2.83} & \textbf{8.37} & \textbf{37.40} & \textbf{1.75} & \textbf{6.42} & 54.50 & 4.25 & \textbf{5.38} & 63.13 & 4.08 & \textbf{12.47} & \textbf{24.01} & \textbf{1.75} & \textbf{2.79} \\
\bottomrule

\end{tabular}
}

\end{table*}

\paragraph{Metrics.}
For every prediction we report three levels of geometric accuracy: scale-invariant (align $s$), affine-invariant (align $s,\mathbf{t}$), and metric (no alignment). Each level is reported both on the recovered point cloud (\emph{Point}) and on the depth channel (\emph{Depth}). Within each level we report AbsRel (denoted Rel) and $\delta^{1.03}$ (the fraction of pixels whose error ratio is below $1.03$), following the MoGe family~\cite{wang2025moge,wang2026moge2}. We also report the per-column rank (\rank) across datasets, and a global Avg. \rank{} obtained by averaging the ranks over all dataset-metric pairs, where Rel and $\delta$ are ranked independently for each dataset. When a method does not produce a particular geometric output (e.g., VideoDA does not regress an absolute-scale 3D point cloud, and deriving depth from inverse disparity results in excessively high Rel errors), the corresponding cell is marked ``\,-\,'' and excluded from rank aggregation.
\rev{In performance tables, the best and second-best displayed values in each metric column are marked in bold and underlined, respectively; displayed ties share the same marking. Rel and rank are lower-is-better, whereas $\delta^{1.03}$ is higher-is-better.}

\paragraph{Baselines.}
We compare against two families. The first are state-of-the-art monocular geometry estimators that operate without test-time adaptation: MoGe2~\cite{wang2026moge2}, UniDepthV2~\cite{piccinelli2025unidepthv2}, and DA3~\cite{lin2025da3}. The second are their \emph{oracle} variants, marked with a superscript $^{*}$. For sequence benchmarks we additionally include three multi-view and one video-depth baseline: VGGT~\cite{wang2025vggt}, DA3-Nested~\cite{lin2025da3}, MapAnything~\cite{keetha2025mapanything}, and VideoDA~\cite{chen2025videoda}. All baselines use their publicly released checkpoints.
\rev{The sequence comparison groups rows into video-depth, single-image, multi-view, oracle-intrinsics, and \octrev{adapted geometry estimation methods}.}

\begin{figure*}[!t]
\input{figure/exp_viz}
\caption{\rev{Qualitative sequence results: metric depth errors and point clouds. Depth errors use a log-scaled plasma color map \raisebox{-0.4ex}{\includegraphics[height=1.6ex,width=3em]{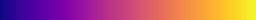}}~(low~$\rightarrow$~high), with a shared range across methods for each sample.}}
\label{fig:viz}
\end{figure*}

\begin{figure*}[!t]
\centering
\begin{minipage}[t]{0.48\textwidth}
\centering
\includegraphics[trim=0cm 0.75cm 0cm 0.7cm,clip,width=\linewidth]{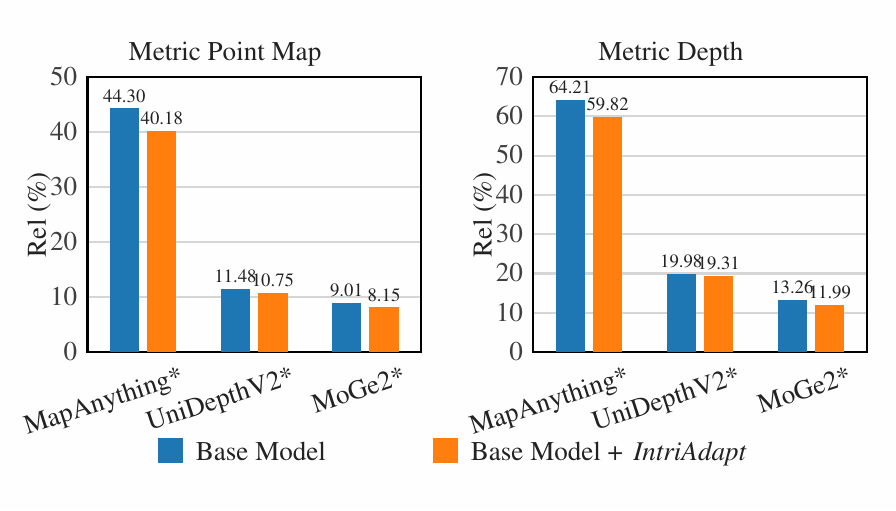}
\caption{\rev{Adaptation across three base models.}}
\label{fig:multi_baseline_rel}
\end{minipage}\hfill
\begin{minipage}[t]{0.48\textwidth}
\centering
\includegraphics[trim=0cm 1.8cm 0cm 0.7cm,clip,width=\linewidth]{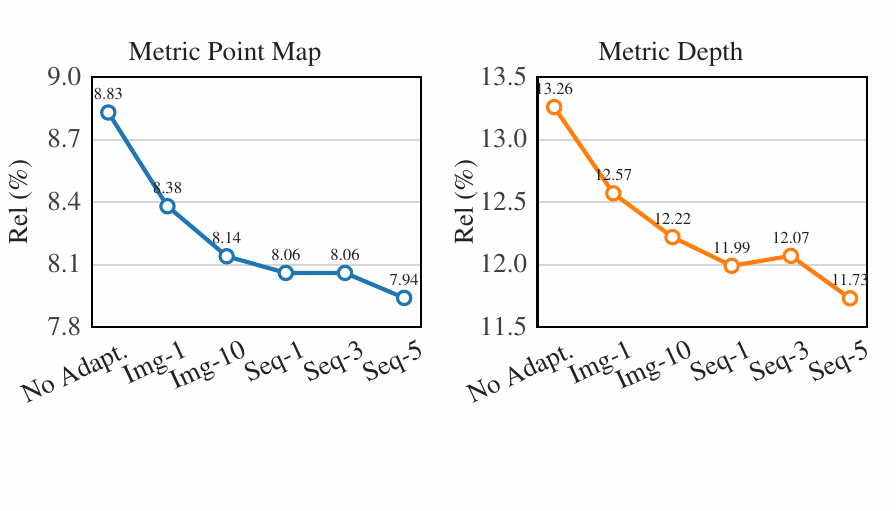}
\caption{\rev{Effect of adaptation data amount and form.}}
\label{fig:abl_samples}
\end{minipage}
\end{figure*}

\subsection{Single-Image Adaptation}
\label{sec:exp_single}

\begin{revision}
Tab.~\ref{tab:sf_main} reports the $10$-dataset averages. Ours$^{*}$ achieves the best overall rank and improves both point-map and depth metrics over MoGe2 and MoGe2$^{*}$. The limited gain from MoGe2 to MoGe2$^{*}$ indicates that the post-hoc solver can only re-assign $(\tau^{*},f^{*})$ on top of an already biased shape. In contrast, Ours$^{*}$ closes the intrinsic loop \emph{inside} the network and reshapes the prediction itself.
\end{revision}

\par\csname @IEEEtranneedspace\endcsname{4\baselineskip}{\relax}
\subsection{Sequence Adaptation}
\label{sec:exp_seq}

\begin{revision}
Tab.~\ref{tab:seq6_main} summarizes the six leakage-free benchmarks. Ours$^{*}$ has the best overall rank ($2.79$), with metric Point and Depth Rel of $8.37$ and $12.47$, compared with $9.25$ and $13.92$ for MoGe2$^{*}$. Its scale- and affine-invariant Rel scores also improve, suggesting that adaptation changes relative geometry as well as metric calibration. Some threshold-accuracy scores still favor other baselines. Because evaluation uses held-out sequences without further updates, the aggregate improvement supports transfer of the learned camera-specific correction beyond the frames used for adaptation. Qualitatively, Fig.~\ref{fig:viz} shows representative results. Each sample contains a ground-truth column followed by five methods. The upper row shows metric depth error, with the input RGB in the ground-truth column; the lower row shows rendered metric point clouds. Ours$^{*}$ reduces the field-of-view distortion visible in MoGe2$^{*}$ while preserving details from the pretrained prediction.
\end{revision}

\begin{table}[!t]
\centering
\small
\setlength{\tabcolsep}{4pt}
\caption{\rev{Temporal consistency: OPW AbsRel ($\downarrow$).}}
\label{tab:opw_temporal}
\begin{tabular}{lcccc}
\toprule
Method & 7-Scenes & ScanNet & Metropolis & Avg. \\
\midrule
VideoDA       & \textbf{0.240} & 0.244 & 0.522 & 0.335 \\
MoGe2         & 0.350 & \underline{0.226} & \underline{0.382} & 0.319 \\
MoGe2$^{*}$   & \underline{0.252} & \underline{0.226} & 0.398 & \underline{0.292} \\
Ours$^{*}$    & 0.260 & \textbf{0.225} & \textbf{0.338} & \textbf{0.274} \\
\bottomrule
\end{tabular}

\end{table}

\begin{revision}
We further evaluate temporal consistency using the optical-flow warping error (OPW)~\cite{wang2022lessismore}. We report OPW on the three consecutive-frame video benchmarks in our suite: 7-Scenes, ScanNet, and Metropolis. As shown in Tab.~\ref{tab:opw_temporal}, adapting the geometric representation does not compromise frame-to-frame consistency \octrev{on the evaluated video sequences}.
\end{revision}


\begin{figure*}[!t]
\centering
\begin{minipage}[t]{0.48\textwidth}
\vspace{0pt}
\centering
\small
\setlength{\tabcolsep}{2pt}
\captionof{table}{\rev{Final focal/shift solver residual ($\downarrow$).}}
\label{tab:opt_loss}
\resizebox{\linewidth}{!}{%
\begin{tabular}{l|ccccccc}
\toprule
Method & 7-Scenes & ScanNet & ETH3D & KITTI & Metro. & DIODE & HAMMER \\
\midrule
MoGe2$^{*}$ & \underline{1.40} & \underline{0.72} & \underline{0.35} & \underline{7.98} & \underline{3.86} & \underline{2.56} & \underline{2.25} \\
Ours$^{*}$ & \textbf{0.30} & \textbf{0.46} & \textbf{0.34} & \textbf{0.92} & \textbf{1.18} & \textbf{0.79} & \textbf{0.81} \\
\bottomrule
\end{tabular}

}

\par\vspace{4pt}
\captionof{table}{\rev{Inference memory and runtime on KITTI.}}
\label{tab:runtime_memory}
\begin{tabular*}{\linewidth}{@{\extracolsep{\fill}}lccc@{}}
\toprule
Method & \shortstack{Peak memory\\(GB)$\downarrow$} & \shortstack{Inference\\(s/frame)$\downarrow$} & \shortstack{TTA\\(s)$\downarrow$} \\
\midrule
MoGe2       & \textbf{2.58} & \textbf{0.24} & -- \\
MoGe2$^{*}$ & \textbf{2.58} & \textbf{0.24} & -- \\
Ours$^{*}$  & \underline{2.64} & \textbf{0.24} & 117.23 \\
\bottomrule
\end{tabular*}

\end{minipage}\hfill
\begin{minipage}[t]{0.48\textwidth}
\vspace{0pt}
\centering
\small
\setlength{\tabcolsep}{2.5pt}
\captionof{table}{\rev{Effect of output-time reprojection (Proj).}}
\label{tab:abl_reproj}
\begin{tabular}{l|cc|cc|cc}
\toprule
\multirow{3}{*}{Method} & \multicolumn{6}{c}{\textit{Point}} \\
\cmidrule(lr){2-7}
 & \multicolumn{2}{c|}{Scale-inv.} & \multicolumn{2}{c|}{Affine-inv.} & \multicolumn{2}{c}{Metric} \\
 & Rel$\downarrow$ & $\delta^{1.03}\!\uparrow$ & Rel$\downarrow$ & $\delta^{1.03}\!\uparrow$ & Rel$\downarrow$ & $\delta^{1.03}\!\uparrow$ \\
\midrule
MoGe2 (w/o Proj) & 9.58 & 18.89 & 6.88 & 47.61 & 9.01 & 30.07 \\
MoGe2 & 9.61 & 18.63 & 6.90 & 47.43 & 9.03 & 30.03 \\
\midrule
MoGe2$^{*}$ (w/o Proj) & 7.39 & 45.38 & 6.88 & 47.61 & 9.01 & 30.07 \\
MoGe2$^{*}$ & 7.03 & 50.85 & 6.54 & 50.95 & 8.83 & 32.71 \\
\midrule
Ours$^{*}$ (w/o Proj) & \underline{6.41} & \underline{53.21} & \underline{6.07} & \underline{54.57} & \underline{8.15} & \underline{36.82} \\
Ours$^{*}$ & \textbf{6.24} & \textbf{55.57} & \textbf{5.94} & \textbf{56.08} & \textbf{8.06} & \textbf{38.48} \\
\bottomrule
\end{tabular}

\end{minipage}
\par\vspace{10pt}
\input{figure/exp_raymap_all}%
\caption{\rev{Ray-map residuals across seven sequence benchmarks, comparing three oracle baselines with \textit{IntriAdapt}.}}
\label{fig:raymap_all}
\end{figure*}

\subsection{Method Analysis}
\label{sec:ablation}

\begin{revision}
We examine generality, data requirements, cost, and the role of each geometric constraint. Unless stated otherwise, the following accuracy ablations average over all seven sequence benchmarks, including ScanNet; they diagnose adaptation behavior and complement the six-dataset comparison above.

\paragraph{Generality across base models.}
Fig.~\ref{fig:multi_baseline_rel} reports metric Point and Depth Rel before and after applying the same intrinsic-guided objective to MapAnything, UniDepthV2, and MoGe2. Metric Point Rel decreases from $44.30$ to $40.18$, $11.48$ to $10.75$, and $9.01$ to $8.15$, respectively. The corresponding metric Depth Rel decreases from $64.21$ to $59.82$, $19.98$ to $19.31$, and $13.26$ to $11.99$. These averages show that encoder adaptation can benefit different prediction heads, including a multi-view model \octrev{with jointly processed image inputs}.

\paragraph{Adaptation data: images vs.\ sequences.}
Fig.~\ref{fig:abl_samples} varies both the quantity and form of target-camera data. Img-$k$ uses $k$ random images, whereas Seq-$k$ uses $k$ complete adaptation sequences and evaluates on held-out sequences from the same camera. Moving from no adaptation to Img-$1$ reduces metric Point/Depth Rel from $8.83/13.26$ to $8.38/12.57$; Img-$10$ reaches $8.14/12.22$, and Seq-$1$ reaches $8.06/11.99$. A single image therefore supplies a useful intrinsic signal, while broader scene coverage strengthens transfer. Beyond Seq-$1$, the additional gain is smaller: Seq-$5$ reaches $7.94/11.73$, improvements of about $1.5\%$ and $2.2\%$ relative to Seq-$1$. This supports Seq-$1$ as a practical default rather than establishing that it is optimal for every metric. The supplementary frame-coverage ablation further shows that $50\%$ of frames per update with $20$ updates gives the best average rank among the tested schedules; increasing the step count alone does not ensure a better result. \fillrev{The shared adapter is reused throughout \octrev{evaluation on previously unseen sequences}.}
\end{revision}

\begin{figure*}[!t]
\centering
\begingroup
\captionof{table}{\rev{Direct point-map optimization versus LoRA adaptation on seven sequence benchmarks. \octrev{All methods use the same oracle intrinsics and final output-time reprojection step}.}}
\label{tab:direct_point_optimization_appendix}
{\small\setlength{\tabcolsep}{2.5pt}\resizebox{\textwidth}{!}{%
\begin{tabular}{l|ccc|ccc|ccc|ccc|ccc|ccc|c}
\toprule
\multirow{3}{*}{Method} & \multicolumn{9}{c|}{\textit{Point}} & \multicolumn{9}{c|}{\textit{Depth}} & Avg. \\
\cmidrule(lr){2-10}\cmidrule(lr){11-19}
 & \multicolumn{3}{c|}{Scale-inv.} & \multicolumn{3}{c|}{Affine-inv.} & \multicolumn{3}{c|}{Metric} & \multicolumn{3}{c|}{Scale-inv.} & \multicolumn{3}{c|}{Affine-inv.} & \multicolumn{3}{c|}{Metric} & Rk.$\downarrow$ \\
 & Rel$\downarrow$ & $\delta^{1.03}\!\uparrow$ & Rk.$\downarrow$ & Rel$\downarrow$ & $\delta^{1.03}\!\uparrow$ & Rk.$\downarrow$ & Rel$\downarrow$ & $\delta^{1.03}\!\uparrow$ & Rk.$\downarrow$ & Rel$\downarrow$ & $\delta^{1.03}\!\uparrow$ & Rk.$\downarrow$ & Rel$\downarrow$ & $\delta^{1.03}\!\uparrow$ & Rk.$\downarrow$ & Rel$\downarrow$ & $\delta^{1.03}\!\uparrow$ & Rk.$\downarrow$ & \\
\midrule
MoGe2$^{*}$ & \underline{7.16} & 50.09 & 2.71 & 6.64 & 50.19 & 2.71 & \underline{9.10} & \underline{32.09} & \underline{2.57} & \underline{7.14} & 50.19 & 2.71 & 5.46 & 62.94 & 2.93 & \underline{13.72} & \underline{20.73} & \underline{2.86} & 2.75 \\
\midrule
MoGe2$^{*}$ Direct-10 (lr=3e-3) & \underline{7.16} & \underline{50.42} & \underline{2.43} & \underline{6.63} & \underline{50.46} & \underline{2.50} & 9.15 & 31.51 & 2.79 & \underline{7.14} & \underline{50.53} & \underline{2.43} & \underline{5.45} & \textbf{63.14} & \textbf{2.14} & 14.01 & 19.84 & 3.21 & \underline{2.58} \\
MoGe2$^{*}$ Direct-10 (lr=3e-1) & 11.63 & 32.23 & 4.93 & 10.91 & 29.63 & 5.00 & 24.29 & 13.64 & 4.79 & 11.60 & 32.33 & 4.93 & 8.54 & 44.29 & 5.00 & 57.81 & 9.67 & 4.57 & 4.87 \\
MoGe2$^{*}$ Direct-10+z & 7.17 & 50.30 & 3.14 & \underline{6.63} & 50.33 & 3.21 & 9.15 & 31.46 & 3.64 & 7.15 & 50.41 & 3.14 & \underline{5.45} & \underline{63.13} & 2.50 & 14.01 & 19.87 & 3.21 & 3.14 \\
\midrule
Ours$^{*}$ & \textbf{6.36} & \textbf{54.86} & \textbf{1.79} & \textbf{6.04} & \textbf{55.30} & \textbf{1.57} & \textbf{8.36} & \textbf{37.54} & \textbf{1.21} & \textbf{6.36} & \textbf{54.91} & \textbf{1.79} & \textbf{5.28} & 63.05 & \underline{2.43} & \textbf{12.56} & \textbf{24.19} & \textbf{1.14} & \textbf{1.65} \\
\bottomrule
\end{tabular}
}
}
\endgroup
\par\vspace{8pt}
\input{figure/appendix_direct_point_optimization_viz}%
\par
\caption{\camrev{Direct point-map optimization and LoRA adaptation. Direct optimization can reduce ray-angle error (lower rows) without improving metric depth (upper rows); IntriAdapt reduces both through ray--depth coupling in model space.}}
\label{fig:direct_point_optimization_viz}
\end{figure*}

\begin{revision}
\paragraph{Why angular supervision improves metric geometry.}
Although the ray-map residual directly constrains angular consistency rather than depth, it improves metric geometry for two reasons. First, aligning predicted rays with the known intrinsics corrects the shape of the point cloud and thereby improves its relative geometry. Second, modern metric geometry models couple ray direction and depth in their learned geometric representations. Correcting ray-map inconsistency therefore updates the shared representation and reshapes the point cloud as a whole, improving metric geometry rather than merely rotating individual rays. Tab.~\ref{tab:opt_loss} shows that adaptation consistently reduces the final focal/shift residual across all sequence benchmarks, indicating improved geometry representation and consistency with the underlying camera model. \fillrev{The largest absolute residual reduction occurs on KITTI, from $7.98$ to $0.92$; on 7-Scenes, the residual falls from $1.40$ to $0.30$. These correspond to relative reductions of $88.5\%$ and $78.6\%$, respectively.} The per-dataset ray-map visualizations in Fig.~\ref{fig:raymap_all} further reinforce this result.

\paragraph{Runtime and memory.}
We profile the KITTI Seq-$1$ setting with MoGe2 in FP32, $3600$ input tokens, and an A100 GPU. Tab.~\ref{tab:runtime_memory} reports $0.24$ seconds per frame for all three variants at the displayed precision. Peak inference memory increases from $2.58$ to $2.64$ GB, a $0.06$ GB increase, while the one-time adaptation stage takes $117.23$ seconds. This separates deployment cost into camera-specific preparation and repeated inference: for $N$ subsequent frames, the measured costs imply approximately $0.24+117.23/N$ seconds per frame including adaptation, ignoring other pipeline overhead. The advantage of sharing an adapter therefore grows when many frames reuse the same camera calibration.

\paragraph{Output-time reprojection.}
Tab.~\ref{tab:abl_reproj} isolates the final projection step. Given the post-shift depth, reprojection fixes each point's viewing direction using the camera ray while keeping its depth coordinate unchanged; consequently, only Point metrics vary. For MoGe2$^{*}$, metric Point Rel decreases from $9.01$ to $8.83$ and $\delta^{1.03}$ rises from $30.07$ to $32.71$. Adaptation already improves these scores to $8.15$ and $36.82$ without reprojection; applying reprojection afterward reaches $8.06$ and $38.48$. The larger gain from adapting the unprojected geometry shows that the benefit is not explained by the final projection alone. Conversely, the remaining improvement after adaptation motivates composing the two operations. The non-oracle MoGe2 rows show a slight degradation after projection with estimated intrinsics, emphasizing that the refinement depends on calibration quality. Reprojection needs no learned update, but it cannot repair an incorrect depth value: its role is to enforce camera consistency on the geometry produced by the model. \fillrev{Relative to the oracle baseline without reprojection, adaptation alone reduces metric Point Rel by $9.5\%$, whereas reprojection alone reduces it by $2.0\%$. Combining both reaches a $10.5\%$ reduction, supporting their complementary roles. These relative gains use the displayed averages and the same $9.01$ reference value.}

\paragraph{Direct point optimization baseline.}
Table~\ref{tab:direct_point_optimization_appendix} compares our LoRA-based test-time adaptation with direct point-map optimization variants. All rows use ground-truth intrinsics at test time. For a fair per-frame optimization budget, the direct baselines use $10$ point-map optimization steps, matching the step count used by our single-image TTA setting. We report both the stable learning rate used in the main direct baseline, $3{\times}10^{-3}$, and a higher learning rate, $3{\times}10^{-1}$, that reduces ray-angle residuals more aggressively. Direct-$10$+z adds a $z$-coordinate retention penalty to the lower-rate configuration to test whether preserving the initial depth stabilizes the optimization. After direct optimization, we apply the same output-time reprojection used by MoGe2$^{*}$ and Ours$^{*}$.

At the lower learning rate, direct optimization changes metric Point Rel from $9.10$ to $9.15$ and metric Depth Rel from $13.72$ to $14.01$. Although some aligned scores improve slightly, absolute geometry does not. Raising the learning rate worsens the two metric Rel scores to $24.29$ and $57.81$. Adding the retention penalty leaves them at $9.15/14.01$, suggesting that constraining changes to the initial depth does not supply the missing depth correction. LoRA adaptation instead reaches $8.36/12.56$ and the best average rank ($1.65$). Even under this favorable setting, directly optimizing the output point map does not match LoRA-based adaptation: aggressive point-level fitting can reduce ray-angle residuals, but it distorts metric geometry and cannot correct the biased depth values. Fig.~\ref{fig:direct_point_optimization_viz} shows the visual comparison. Each sample shows RGB, MoGe2$^{*}$, the two Direct-$10$ learning rates, and Ours$^{*}$. Upper rows contain metric depth errors using the shared per-sample log-plasma range, while lower rows contain polar-angle errors, with the RGB column left blank. \camrev{Annotations report mean absolute error (MAE) in meters for depth and degrees for ray angle.} The angular errors are computed before final reprojection, which would otherwise remove the residual by construction.
\end{revision}

\raggedbottom
\section{Conclusion}
\label{sec:conclusion}



This paper shows that camera intrinsics can do more than condition a feed-forward metric geometry model: they can verify its output. By turning the mismatch between predicted and camera-defined ray maps into a test-time signal, \textit{IntriAdapt} uses information already available in calibrated cameras to refine metric geometry without depth labels. The consistent gains over both unadapted and oracle-intrinsics baselines suggest that simply providing intrinsics to a model is not enough. Using them to close the output loop is crucial.
Beyond per-image adaptation, our results indicate that the learned correction is largely camera-level. An adapter trained on a small representative sequence transfers to held-out frames from the same sensor, making intrinsic-guided adaptation a \octrev{reusable calibration mechanism for that camera}.

\bibliographystyle{IEEEtran}
\bibliography{intriadapt}

\onecolumn
\newgeometry{left=0.696in,right=0.929in,top=1in,textheight=8.875in,headheight=0pt,headsep=0pt,footskip=30pt,twocolumn=false}
\pagestyle{plain}
\markboth{}{}
\raggedbottom
\setlength{\parindent}{1pc}
\hfuzz=0.1pt
\vfuzz=0.1pt
\emergencystretch=1em
\setcounter{topnumber}{4}
\setcounter{bottomnumber}{2}
\setcounter{totalnumber}{6}
\renewcommand{\topfraction}{0.99}
\renewcommand{\bottomfraction}{0.99}
\renewcommand{\textfraction}{0.01}
\renewcommand{\floatpagefraction}{0.90}
\setcounter{section}{0}
\setcounter{subsection}{0}
\setcounter{table}{0}
\setcounter{figure}{0}
\setcounter{equation}{0}
\renewcommand{\thesection}{\arabic{section}}
\renewcommand{\thesubsection}{\thesection.\arabic{subsection}}
\renewcommand{\thetable}{S\arabic{table}}
\renewcommand{\thefigure}{S\arabic{figure}}
\renewcommand{\theequation}{S\arabic{equation}}
\renewcommand{\theHsection}{supp.\arabic{section}}
\renewcommand{\theHsubsection}{\theHsection.\arabic{subsection}}
\renewcommand{\theHtable}{supp.\arabic{table}}
\renewcommand{\theHfigure}{supp.\arabic{figure}}
\renewcommand{\theHequation}{supp.\arabic{equation}}
\captionsetup{font=small}
\captionsetup[table]{name=Table,labelsep=period,aboveskip=3pt,belowskip=2pt}
\captionsetup[figure]{name=Fig.,labelsep=period,aboveskip=5pt,belowskip=0pt}
\newsavebox{\detailtablebox}
\newcommand{\fitdetailtable}[1]{%
  \sbox{\detailtablebox}{\resizebox{\linewidth}{!}{#1}}%
  \ifdim\dimexpr\ht\detailtablebox+\dp\detailtablebox\relax>\dimexpr\textheight-25pt\relax
    \resizebox*{!}{\dimexpr\textheight-25pt\relax}{\usebox{\detailtablebox}}%
  \else
    \usebox{\detailtablebox}%
  \fi}
\makeatletter
\setlength{\@fptop}{0pt}
\renewcommand{\@seccntformat}[1]{\csname the#1\endcsname.\quad}
\renewcommand{\section}{\@startsection{section}{1}{\z@}{10pt plus 2pt minus 2pt}{7pt}{\large\bfseries}}
\renewcommand{\subsection}{\@startsection{subsection}{2}{\z@}{8pt plus 2pt minus 2pt}{6pt}{\fontsize{11}{13}\selectfont\bfseries}}
\renewcommand{\l@section}[2]{%
  \ifnum\c@tocdepth>\z@
    \addpenalty\@secpenalty\addvspace{1em plus 1pt}%
    \begingroup\parindent\z@\rightskip\@pnumwidth
      \parfillskip-\@pnumwidth\leavevmode\bfseries
      \advance\leftskip1.5em\hskip-\leftskip
      #1\nobreak\hfil\nobreak\hb@xt@\@pnumwidth{\hss #2}\par
    \endgroup
  \fi}
\renewcommand{\l@subsection}{\@dottedtocline{2}{1.5em}{2.3em}}
\makeatother
\addtocontents{toc}{\protect\setcounter{tocdepth}{2}}
\phantomsection
\pdfbookmark[0]{Supplementary Material}{supplementary}
\null\vskip .375in
\begin{center}
{\Large\bfseries Supplementary Material for \emph{IntriAdapt}\par}
\vspace*{24pt}
{\large Jiaxing Chen, Zizheng Yan, Jingyu Zhuang, Jinwei Chen,\\
Shuguang Cui, Guanying Chen, Qingnan Fan\par}
\vskip .5em
\vspace*{12pt}
\end{center}
\makeatletter\global\@topnum\z@\makeatother
\vspace{-12pt}
{\hypersetup{linkcolor=black}\tableofcontents}
\let\appendix\relax

\appendix

\section{More Details of the Datasets}
\label{sec:app_data}

\subsection{Single-Image Benchmark FoV Distribution}
\label{sec:app_data_singleframe_fov}

The $10$ single-image benchmarks adopted from the MoGe2~\octrev{\cite{wang2026moge2}} protocol cover a broad range of intrinsic regimes. We summarize the average GT FoV of each benchmark in Table~\ref{tab:app_singleframe_fov}. We report the FoV \emph{after} the benchmark preprocessing used by the MoGe2~\octrev{\cite{wang2026moge2}} evaluator, i.e., after projecting each image to the target resolution listed in the table. \rev{Rows are sorted by mean processed horizontal FoV.}

\subsection{Sequence Benchmark Construction}
\label{sec:app_seq_construction}

Our 7 sequence benchmarks combine three densely-captured video datasets adopted by CAPA~\cite{ke2026capa} (7-Scenes, ScanNet, Metropolis) with four high-diversity single-image benchmarks (ETH3D, KITTI, DIODE, HAMMER) re-organized into sensor-indexed sequences. The motivation for re-purposing single-image datasets is precisely that dense videos exhibit very small inter-frame variability and therefore probe sensor-level generalization weakly; sequences mined from a static-camera image collection have much larger inter-frame baseline and content diversity while still sharing one physical sensor.

\paragraph{Construction protocol.}
A valid sequence dataset, in our setup, must satisfy two intrinsic-consistency requirements: (i) all images inside one sequence share a single $3{\times}3$ intrinsic matrix, and (ii) all sequences inside one dataset that we average over also share that intrinsic matrix. For every candidate dataset we therefore group images by their per-image intrinsic matrix and accept a group as one sequence only if all members agree within an absolute tolerance of $\mathrm{atol}{=}0.005$ on the normalized intrinsic entries. \rev{We chose $0.005$ because it keeps the reported within-group FoV deviation at or below $0.57^\circ$ on all datasets (Table~\ref{tab:app_atol_fov_residual}). The mean FoV columns use the processed target view of the evaluator.} Looser tolerances ($\mathrm{atol}{=}0.05$) merge cameras that differ by several degrees of FoV and therefore violate requirement (ii); tighter tolerances ($\mathrm{atol}{=}0.0005$) over-fragment the data and shrink the largest usable camera group to a few sequences only.

\begin{table}[!t]
\centering
\begin{minipage}[t]{0.38\textwidth}
\vspace{0pt}
\centering
\small
\setlength{\tabcolsep}{2.5pt}
\caption{\rev{\octrev{Horizontal and vertical FoV of the ten processed single-image benchmarks.}}}
\label{tab:app_singleframe_fov}
\begin{tabular*}{\linewidth}{@{\extracolsep{\fill}}l|c|cc@{}}
\toprule
Dataset & Resolution & FoVx ($^\circ$) & FoVy ($^\circ$) \\
\midrule
DIODE   & $1024{\times}768$  & 42.3 & 32.4 \\
Spring  & $1920{\times}1080$ & 44.7 & 26.4 \\
GSO     & $512{\times}512$   & 45.1 & 45.1 \\
Sintel  & $872{\times}436$   & 51.1 & 27.3 \\
KITTI   & $750{\times}375$   & 54.4 & 28.8 \\
iBims-1 & $640{\times}480$   & 56.7 & 44.1 \\
NYUv2   & $640{\times}480$   & 60.6 & 47.3 \\
HAMMER  & $1664{\times}832$  & 63.7 & 34.5 \\
DDAD    & $1400{\times}700$  & 70.7 & 39.5 \\
ETH3D   & $2048{\times}1365$ & 84.3 & 62.2 \\
\bottomrule
\end{tabular*}
\end{minipage}%
\hfill
\begin{minipage}[t]{0.60\textwidth}
\vspace{0pt}
\centering
\small
\setlength{\tabcolsep}{2pt}
\caption{\rev{FoV deviation at three intrinsic-grouping tolerances.}}
\label{tab:app_atol_fov_residual}
\begin{tabular*}{\linewidth}{@{\extracolsep{\fill}}l|cc|cc|cc|cc@{}}
\toprule
\multirow{2}{*}{Dataset} & \multicolumn{2}{c|}{\shortstack{Mean FoV\\($^\circ$)}} & \multicolumn{2}{c|}{\shortstack{$\Delta$FoV\\@ $0.05$}} & \multicolumn{2}{c|}{\shortstack{$\Delta$FoV\\@ $0.005$}} & \multicolumn{2}{c}{\shortstack{$\Delta$FoV\\@ $0.0005$}} \\
 & x & y & x & y & x & y & x & y \\
\midrule
DIODE       & 42.3 & 32.4 & 1.45 & 0.87 & 0.15 & 0.09 & 0.015 & 0.009 \\
KITTI       & 54.3 & 28.8 & 4.64 & 0.71 & 0.49 & 0.07 & 0.049 & 0.007 \\
7-Scenes    & 57.4 & 44.6 & 2.53 & 1.59 & 0.26 & 0.16 & 0.026 & 0.016 \\
ScanNet     & 57.9 & 45.0 & 2.60 & 1.63 & 0.27 & 0.17 & 0.027 & 0.017 \\
HAMMER      & 63.7 & 34.5 & 3.61 & 1.01 & 0.38 & 0.10 & 0.038 & 0.010 \\
ETH3D       & 84.3 & 62.2 & 4.93 & 2.95 & 0.52 & 0.31 & 0.052 & 0.031 \\
Metropolis  & 90.0 & 90.0 & 5.45 & 5.45 & 0.57 & 0.57 & 0.057 & 0.057 \\
\bottomrule
\end{tabular*}
\end{minipage}
\end{table}

\begin{figure}[!p]
\centering
\captionsetup{name=Fig.}
\ifdatasetreview\captionsetup{labelfont={color=blue}}\fi
\input{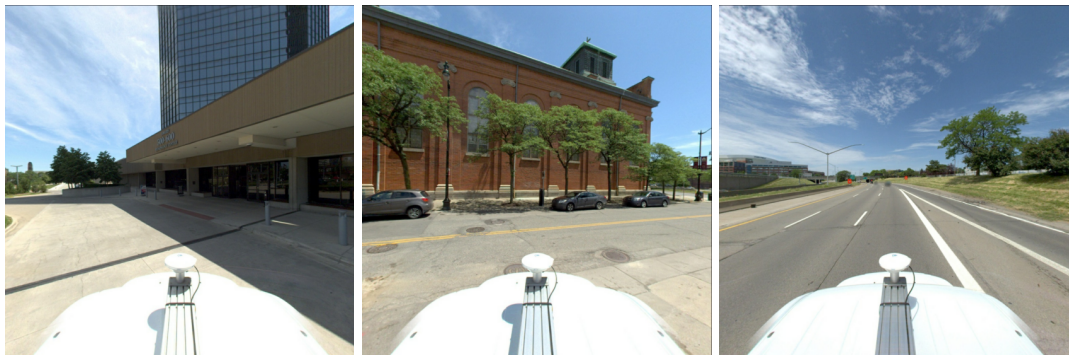}
\caption{\datasetrev{Diversity of the sequence benchmark. Each row shows three different sequences from one retained intrinsic group. Image aspect ratios follow evaluation. The examples cover indoor rooms, driving scenes, outdoor structures, and household materials. They retain the evaluation preprocessing, revealing scene variation within intrinsic groups and camera FoV differences across groups.}}
\label{fig:app_dataset_diversity}
\end{figure}

\paragraph{Benchmark composition and statistics.}
Under this filter, the four re-organized single-image datasets (KITTI\_09\_26, ETH3D, DIODE, and HAMMER) and Metropolis form intrinsic-coherent groups in the current evaluation index. For KITTI, we use the KITTI\_09\_26 camera group rather than the full benchmark, which contains multiple intrinsic groups. ETH3D requires light cleaning: we discard the \texttt{office} sequence because its photometric calibration yields a different intrinsic matrix, as well as a small number of frames in \texttt{pipes} whose intrinsics deviate beyond $0.005$. Since 7-Scenes and ScanNet contain multiple camera groups by design, we retain the largest intrinsic-coherent group in each prepared evaluation split. For every selected dataset, we cap the sequence length at the dense-video subsampling rate adopted by CAPA~\octrev{\cite{ke2026capa}} for 7-Scenes, ScanNet, and Metropolis, or at approximately $25$--$40$ frames for the re-organized image collections. Table~\ref{tab:app_seq_summary} summarizes the resulting benchmark, including its processed FoV, number of sequences, average sequence length, and total number of frames. \rev{Rows are sorted by processed horizontal FoV; the main paper includes a compact copy without the identifier, scene type, and resolution columns.} \datasetrev{Fig.~\ref{fig:app_dataset_diversity} illustrates this diversity.}

\begin{table*}[t]
\centering
\small
\setlength{\tabcolsep}{4pt}
\caption{\rev{Statistics of the seven sequence benchmarks after intrinsic grouping.}}
\label{tab:app_seq_summary}
\begin{tabular}{c|l|l|c|cc|c|c|c}
\toprule
\# & Dataset & Type & Resolution & FoVx ($^\circ$) & FoVy ($^\circ$) & \#Seqs & Frames/Seq & \#Frames \\
\midrule
1 & DIODE       & Indoor + outdoor       & $1024{\times}768$  & 42.3 & 32.4 & 20 & 38.5  & 771 \\
2 & KITTI\_09\_26 & Driving              & $750{\times}375$   & 54.3 & 28.8 & 21 & 23.2  & 487 \\
3 & 7-Scenes    & Indoor                 & $640{\times}480$   & 57.4 & 44.6 & 18 & 100.0 & 1800 \\
4 & ScanNet     & Indoor                 & $640{\times}480$   & 57.9 & 45.0 & 56 & 100.0 & 5600 \\
5 & HAMMER      & Multi-sensor handheld  & $1664{\times}832$  & 63.7 & 34.5 & 29 & 26.7  & 775 \\
6 & ETH3D       & Indoor + outdoor       & $2048{\times}1365$ & 84.3 & 62.2 & 12 & 35.7  & 428 \\
7 & Metropolis  & Cityscape              & $1024{\times}1024$ & 90.0 & 90.0 & 35 & 95.2  & 3332 \\
\bottomrule
\end{tabular}
\end{table*}

\section{More Results for Method Analysis}

\rev{The complete implementation details, including the LoRA~\octrev{\cite{hu2022lora}} configuration, per-image and sequence update schedules, frame batching, ray-map resolution, held-out evaluation, and the UniDepthV2~\octrev{\cite{piccinelli2025unidepthv2}} \octrev{and} MapAnything~\octrev{\cite{keetha2025mapanything}} settings, are provided in the main paper's experimental setup. The tables below give the additional ablations.}

\begin{table}[!htbp]
\centering
\small
\setlength{\tabcolsep}{2.5pt}
\caption{\rev{Frame coverage and update count under Seq-$1$ adaptation.}}
\label{tab:abl_pct_step}
\resizebox{\textwidth}{!}{%
\begin{tabular}{ll|ccc|ccc|ccc|ccc|ccc|ccc|c}
\toprule
\multirow{3}{*}{frames [\%]} & \multirow{3}{*}{steps} & \multicolumn{9}{c|}{\textit{Point}} & \multicolumn{9}{c|}{\textit{Depth}} & Avg. \\
\cmidrule(lr){3-11}\cmidrule(lr){12-20}
 &  & \multicolumn{3}{c|}{Scale-inv.} & \multicolumn{3}{c|}{Affine-inv.} & \multicolumn{3}{c|}{Metric} & \multicolumn{3}{c|}{Scale-inv.} & \multicolumn{3}{c|}{Affine-inv.} & \multicolumn{3}{c|}{Metric} & \rank$\downarrow$ \\
 &  & Rel$\downarrow$ & $\delta^{1.03}\!\uparrow$ & \rank$\downarrow$ & Rel$\downarrow$ & $\delta^{1.03}\!\uparrow$ & \rank$\downarrow$ & Rel$\downarrow$ & $\delta^{1.03}\!\uparrow$ & \rank$\downarrow$ & Rel$\downarrow$ & $\delta^{1.03}\!\uparrow$ & \rank$\downarrow$ & Rel$\downarrow$ & $\delta^{1.03}\!\uparrow$ & \rank$\downarrow$ & Rel$\downarrow$ & $\delta^{1.03}\!\uparrow$ & \rank$\downarrow$ & \\
\midrule
100 & 10 & \underline{6.36} & \underline{54.91} & \underline{2.14} & 6.03 & \underline{55.52} & \underline{2.21} & 8.22 & 37.34 & 3.14 & \underline{6.35} & \underline{54.99} & \underline{2.14} & 5.23 & \textbf{64.04} & \textbf{1.86} & 12.27 & 23.81 & 3.29 & \underline{2.46} \\
50 & 20 & \textbf{6.24} & \textbf{55.57} & \textbf{1.71} & \textbf{5.94} & \textbf{56.08} & \textbf{1.79} & \textbf{8.06} & \textbf{38.48} & \textbf{1.64} & \textbf{6.23} & \textbf{55.63} & \textbf{1.71} & \underline{5.20} & \underline{63.98} & \underline{2.00} & \textbf{11.99} & 24.60 & \textbf{2.00} & \textbf{1.81} \\
20 & 50 & \underline{6.36} & 54.28 & 3.00 & \underline{6.02} & 54.81 & 2.86 & \underline{8.13} & \underline{37.73} & 2.64 & \underline{6.35} & 54.33 & 3.00 & \textbf{5.19} & 63.77 & 2.71 & \underline{12.07} & \underline{24.69} & \underline{2.21} & 2.74 \\
10 & 100 & 6.40 & 53.81 & 3.14 & 6.05 & 54.37 & 3.14 & 8.15 & 37.51 & \underline{2.57} & 6.39 & 53.87 & 3.14 & 5.21 & 63.68 & 3.43 & 12.10 & \textbf{24.73} & 2.50 & 2.99 \\
\bottomrule
\end{tabular}
    }

\end{table}

\begin{table}[!htbp]
\centering
\small
\setlength{\tabcolsep}{2.5pt}
\caption{\rev{Adaptation data ablation across seven sequence benchmarks.}}
\label{tab:app_abl_samples}
\resizebox{\textwidth}{!}{%
\begin{tabular}{l|ccc|ccc|ccc|ccc|ccc|ccc|c}
\toprule
\multirow{3}{*}{Adaptation} & \multicolumn{9}{c|}{\textit{Point}} & \multicolumn{9}{c|}{\textit{Depth}} & Avg. \\
\cmidrule(lr){2-10}\cmidrule(lr){11-19}
 & \multicolumn{3}{c|}{Scale-inv.} & \multicolumn{3}{c|}{Affine-inv.} & \multicolumn{3}{c|}{Metric} & \multicolumn{3}{c|}{Scale-inv.} & \multicolumn{3}{c|}{Affine-inv.} & \multicolumn{3}{c|}{Metric} & \rank$\downarrow$ \\
 & Rel$\downarrow$ & $\delta^{1.03}\!\uparrow$ & \rank$\downarrow$ & Rel$\downarrow$ & $\delta^{1.03}\!\uparrow$ & \rank$\downarrow$ & Rel$\downarrow$ & $\delta^{1.03}\!\uparrow$ & \rank$\downarrow$ & Rel$\downarrow$ & $\delta^{1.03}\!\uparrow$ & \rank$\downarrow$ & Rel$\downarrow$ & $\delta^{1.03}\!\uparrow$ & \rank$\downarrow$ & Rel$\downarrow$ & $\delta^{1.03}\!\uparrow$ & \rank$\downarrow$ & \\
\midrule
No adapt. & 7.03 & 50.85 & 5.93 & 6.54 & 50.95 & 6.00 & 8.83 & 32.71 & 5.79 & 7.01 & 50.96 & 5.86 & 5.38 & 63.90 & 3.93 & 13.26 & 20.77 & 5.29 & 5.46 \\
\midrule
Img-1 & 6.50 & 54.43 & 3.93 & 6.15 & 55.03 & 4.21 & 8.38 & 36.30 & 4.64 & 6.49 & 54.52 & 3.86 & 5.32 & \underline{64.05} & \textbf{2.43} & 12.57 & 22.91 & 4.43 & 3.92 \\
Img-10 & 6.28 & 55.41 & 3.50 & 5.98 & 55.92 & 3.50 & 8.14 & 37.37 & 3.64 & 6.27 & 55.47 & 3.50 & \underline{5.23} & 63.97 & \underline{3.43} & 12.22 & 23.41 & 3.57 & 3.52 \\
Seq-$1$ & 6.24 & \underline{55.57} & 3.43 & 5.94 & 56.08 & 3.36 & \underline{8.06} & \underline{38.48} & \underline{2.50} & 6.23 & \underline{55.63} & 3.57 & \textbf{5.20} & 63.98 & 3.71 & \underline{11.99} & \textbf{24.60} & \textbf{2.43} & 3.17 \\
Seq-$3$ & \underline{6.22} & 55.52 & \underline{2.36} & \underline{5.93} & \underline{56.13} & \underline{2.14} & \underline{8.06} & 38.29 & \underline{2.50} & \underline{6.21} & 55.59 & \underline{2.43} & \textbf{5.20} & 64.03 & 3.71 & 12.07 & 23.52 & 2.79 & \underline{2.65} \\
Seq-$5$ & \textbf{6.21} & \textbf{55.80} & \textbf{1.86} & \textbf{5.91} & \textbf{56.41} & \textbf{1.79} & \textbf{7.94} & \textbf{38.90} & \textbf{1.93} & \textbf{6.20} & \textbf{55.86} & \textbf{1.79} & \textbf{5.20} & \textbf{64.14} & 3.79 & \textbf{11.73} & \underline{24.34} & \underline{2.50} & \textbf{2.27} \\

\bottomrule
\end{tabular}

}

\end{table}

\begin{table}[!htbp]
\centering
\small
\setlength{\tabcolsep}{2.5pt}
\caption{\rev{Sequence comparison on all seven benchmarks, including ScanNet.}}
\label{tab:seq7_appendix}
\resizebox{\textwidth}{!}{%
\begin{tabular}{l|ccc|ccc|ccc|ccc|ccc|ccc|c}
\toprule
\multirow{3}{*}{Method} & \multicolumn{9}{c|}{\textit{Point}} & \multicolumn{9}{c|}{\textit{Depth}} & Avg. \\
\cmidrule(lr){2-10}\cmidrule(lr){11-19}
 & \multicolumn{3}{c|}{Scale-inv.} & \multicolumn{3}{c|}{Affine-inv.} & \multicolumn{3}{c|}{Metric} & \multicolumn{3}{c|}{Scale-inv.} & \multicolumn{3}{c|}{Affine-inv.} & \multicolumn{3}{c|}{Metric} & \rank$\downarrow$ \\
 & Rel$\downarrow$ & $\delta^{1.03}\!\uparrow$ & \rank$\downarrow$ & Rel$\downarrow$ & $\delta^{1.03}\!\uparrow$ & \rank$\downarrow$ & Rel$\downarrow$ & $\delta^{1.03}\!\uparrow$ & \rank$\downarrow$ & Rel$\downarrow$ & $\delta^{1.03}\!\uparrow$ & \rank$\downarrow$ & Rel$\downarrow$ & $\delta^{1.03}\!\uparrow$ & \rank$\downarrow$ & Rel$\downarrow$ & $\delta^{1.03}\!\uparrow$ & \rank$\downarrow$ & \\
\midrule
VideoDA & - & - & - & - & - & - & - & - & - & - & 18.91 & 10.86 & - & 36.04 & 10.86 & - & - & - & 10.86 \\
\midrule
DA3 & - & - & - & - & - & - & - & - & - & 7.92 & 49.21 & 7.86 & 7.06 & 57.10 & 7.86 & 14.70 & 19.86 & \underline{3.64} & 6.45 \\
UniDepthV2 & 11.24 & 32.05 & 6.00 & 9.44 & 49.68 & 5.00 & 14.20 & 26.95 & 5.00 & 7.96 & 57.72 & 4.50 & 7.17 & 63.96 & 4.71 & 26.55 & 14.00 & 6.14 & 5.23 \\
MoGe2 & 9.61 & 18.63 & 6.93 & 6.90 & 47.43 & 6.00 & 9.03 & 30.03 & 4.50 & \underline{6.34} & 55.58 & 4.93 & \underline{5.38} & 63.90 & 4.71 & 15.79 & 12.23 & 5.71 & 5.46 \\
\midrule
VGGT & 17.25 & 16.37 & 7.14 & 10.56 & 27.06 & 7.50 & - & - & - & 7.04 & 49.98 & 6.64 & 6.17 & 54.90 & 7.29 & - & - & - & 7.14 \\
MapAnything & 11.29 & 17.11 & 7.21 & 8.19 & 42.34 & 6.93 & 54.79 & 9.58 & 7.07 & 8.17 & 51.43 & 6.71 & 6.33 & 59.07 & 6.86 & 70.91 & 5.33 & 7.93 & 7.12 \\
DA3-Nested & 9.87 & 45.76 & 4.29 & 8.82 & 52.05 & 4.36 & 11.18 & 31.69 & \underline{3.50} & 7.42 & \textbf{60.61} & \underline{4.29} & 6.52 & \textbf{65.97} & \underline{4.43} & 19.45 & 15.82 & 4.00 & 4.14 \\
\midrule
UniDepthV2$^{*}$ & 7.90 & \textbf{56.56} & \textbf{2.07} & 7.48 & \textbf{58.31} & \textbf{2.50} & 11.47 & \underline{33.18} & 4.00 & 7.61 & \underline{58.32} & \textbf{3.36} & 6.94 & 63.93 & \textbf{4.07} & 19.98 & 20.10 & 3.93 & \underline{3.32} \\
MoGe2$^{*}$ & \underline{7.03} & 50.85 & 3.86 & \underline{6.54} & 50.95 & 4.93 & \underline{8.83} & 32.71 & 3.64 & 7.01 & 50.96 & 6.79 & \underline{5.38} & 63.90 & \underline{4.43} & \underline{13.26} & \underline{20.77} & 3.93 & 4.60 \\
MapAnything$^{*}$ & 8.63 & 40.36 & 5.07 & 7.19 & 52.87 & 4.50 & 44.38 & 12.67 & 6.21 & 7.93 & 53.37 & 5.36 & 6.26 & 59.76 & 6.07 & 64.30 & 6.65 & 7.50 & 5.79 \\
\midrule
Ours$^{*}$ & \textbf{6.24} & \underline{55.57} & \underline{2.43} & \textbf{5.94} & \underline{56.08} & \underline{3.29} & \textbf{8.06} & \textbf{38.48} & \textbf{2.07} & \textbf{6.23} & 55.63 & 4.64 & \textbf{5.20} & \underline{63.98} & 4.64 & \textbf{11.99} & \textbf{24.60} & \textbf{2.21} & \textbf{3.21} \\
\bottomrule
\end{tabular}
}

\end{table}

\begin{revision}
\subsection{\octrev{Frame coverage vs.\ step count}}
Table~\ref{tab:abl_pct_step} varies the percentage of adaptation frames per update and the number of AdamW updates under Seq-$1$ with oracle intrinsics. Other settings follow the main-paper implementation, and results average over seven sequence benchmarks. The product of frame coverage and update count is fixed, giving approximately equal frame exposure rather than a measured equality of wall-clock time. The $(50\%,20)$ schedule has the best average rank ($1.81$), compared with $2.46$ for $(100\%,10)$ and $2.99$ for $(10\%,100)$. It also has the lowest metric Point and Depth Rel ($8.06/11.99$), although it does not lead every aligned or threshold metric. In particular, $(20\%,50)$ gives the lowest affine-invariant Depth Rel ($5.19$), while $(10\%,100)$ gives the highest metric Depth $\delta^{1.03}$ ($24.73$). Thus, moderate frame coverage and update count provide the strongest aggregate result in this sweep; neither the largest batch nor the most updates is uniformly preferable.

\subsection{\octrev{Adaptation data granularity}}
Table~\ref{tab:app_abl_samples} gives the complete metric suite for the main-paper data ablation, averaged over seven sequence benchmarks. Img-$k$ uses $k$ random images from the target camera, whereas Seq-$k$ uses $k$ full sequences and evaluates on held-out sequences from the same camera. Seq-$5$ obtains the best average rank ($2.27$), but Seq-$1$ already captures most of the metric Rel reduction and has the highest metric Depth $\delta^{1.03}$ ($24.60$). Seq-$3$ slightly worsens metric Depth Rel relative to Seq-$1$, illustrating that added data does not improve every score monotonically. The default Seq-$1$ setting limits the amount of camera-specific adaptation data while retaining substantial gains.
\end{revision}

\subsection{\octrev{Full $7$-dataset sequence comparison including ScanNet}} Table~\ref{tab:seq7_appendix} repeats the sequence comparison on the full $7$-dataset average, including ScanNet, as a complement to the leakage-free $6$-dataset main result. Because ScanNet is part of the training data of VGGT~\octrev{\cite{wang2025vggt}}, UniDepthV2~\octrev{\cite{piccinelli2025unidepthv2}}, and VideoDA~\octrev{\cite{chen2025videoda}}, the absolute numbers of those rows on this table are optimistic; we include it for completeness so that direct numerical comparison with prior work that reports ScanNet is still possible. The relative ordering of methods, in particular the leading rank of Ours\!$^{*}$, is preserved between the two tables.

\clearpage
\section{Per-Dataset Detailed Results}
\label{sec:app_perdataset}

\subsection{\camrev{Detailed results on single-image and sequence benchmarks}}
\camrev{We separate metric accuracy from scale- and affine-invariant geometry. A superscript $^{*}$ denotes known intrinsics. In Tables~\ref{tab:singleframe_metric}--\ref{tab:sequence_perdataset_delta103}, orange-to-green backgrounds indicate worse-to-better scores; bold and underlining mark the best and next distinct values. Lower Rel/rank and higher $\delta^{1.03}$ are better. Ranks average dataset--metric pairs within each block.}

\par\smallskip
\noindent\begin{minipage}{\linewidth}
\centering
\scriptsize
\setlength{\tabcolsep}{1.6pt}
\renewcommand{\arraystretch}{1.0}
\captionof{table}{\camrev{Single-image metric accuracy on seven benchmarks.}}
\label{tab:singleframe_metric}
\fitdetailtable{%
\begin{tabular}{l|cc|cc|cc|cc|cc|cc|cc|ccc}
\toprule
\multirow{2}{*}{Method} & \multicolumn{2}{c|}{NYUv2} & \multicolumn{2}{c|}{KITTI} & \multicolumn{2}{c|}{ETH3D} & \multicolumn{2}{c|}{iBims-1} & \multicolumn{2}{c|}{DDAD} & \multicolumn{2}{c|}{DIODE} & \multicolumn{2}{c|}{HAMMER} & \multicolumn{3}{c}{Avg.} \\
\cmidrule(lr){2-3}\cmidrule(lr){4-5}\cmidrule(lr){6-7}\cmidrule(lr){8-9}\cmidrule(lr){10-11}\cmidrule(lr){12-13}\cmidrule(lr){14-15}\cmidrule(lr){16-18}
 & Rel$\downarrow$ & $\delta^{1.03}\!\uparrow$ & Rel$\downarrow$ & $\delta^{1.03}\!\uparrow$ & Rel$\downarrow$ & $\delta^{1.03}\!\uparrow$ & Rel$\downarrow$ & $\delta^{1.03}\!\uparrow$ & Rel$\downarrow$ & $\delta^{1.03}\!\uparrow$ & Rel$\downarrow$ & $\delta^{1.03}\!\uparrow$ & Rel$\downarrow$ & $\delta^{1.03}\!\uparrow$ & Rel$\downarrow$ & $\delta^{1.03}\!\uparrow$ & \rank$\downarrow$ \\
\midrule
\multicolumn{18}{c}{\textit{Point map (metric)}} \\
\midrule
DA3 & - & - & - & - & - & - & - & - & - & - & - & - & - & - & - & - & - \\
MapAnything & \cellcolor[rgb]{0.993,0.717,0.409} 9.91 & \cellcolor[rgb]{0.992,0.686,0.384} 14.72 & \cellcolor[rgb]{0.991,0.677,0.378} 11.24 & \cellcolor[rgb]{0.991,0.677,0.378} 14.41 & \cellcolor[rgb]{0.991,0.677,0.378} 11.09 & \cellcolor[rgb]{0.993,0.709,0.403} 13.65 & \cellcolor[rgb]{0.991,0.677,0.378} 8.92 & \cellcolor[rgb]{0.991,0.677,0.378} 24.79 & \cellcolor[rgb]{0.991,0.677,0.378} 17.06 & \cellcolor[rgb]{0.991,0.677,0.378} 12.34 & \cellcolor[rgb]{0.974,0.989,0.713} 11.76 & \cellcolor[rgb]{0.994,0.778,0.461} 19.39 & \cellcolor[rgb]{0.991,0.677,0.378} 44.38 & \cellcolor[rgb]{0.991,0.677,0.378} 0.40 & \cellcolor[rgb]{0.991,0.677,0.378} 16.34 & \cellcolor[rgb]{0.991,0.677,0.378} 14.24 & \cellcolor[rgb]{0.991,0.677,0.378} 6.57 \\
UniDepthV2 & \cellcolor[rgb]{0.741,0.890,0.474} 4.83 & \cellcolor[rgb]{0.788,0.910,0.504} 48.97 & \cellcolor[rgb]{0.757,0.897,0.484} 5.88 & \cellcolor[rgb]{0.827,0.927,0.530} 44.13 & \cellcolor[rgb]{0.997,0.917,0.609} 9.46 & \cellcolor[rgb]{0.996,0.888,0.561} 18.63 & \cellcolor[rgb]{0.773,0.903,0.494} \underline{5.23} & \cellcolor[rgb]{0.909,0.962,0.625} \underline{44.66} & \cellcolor[rgb]{0.950,0.979,0.681} 13.27 & \cellcolor[rgb]{0.999,0.955,0.673} 19.57 & \cellcolor[rgb]{0.991,0.677,0.378} 16.96 & \cellcolor[rgb]{0.991,0.677,0.378} 16.41 & \cellcolor[rgb]{0.725,0.883,0.464} 14.98 & \cellcolor[rgb]{0.996,0.883,0.553} 8.31 & \cellcolor[rgb]{0.886,0.952,0.593} 10.09 & \cellcolor[rgb]{0.962,0.984,0.697} 28.67 & \cellcolor[rgb]{0.999,0.969,0.697} 4.36 \\
MoGe2 & \cellcolor[rgb]{0.655,0.853,0.418} 4.21 & \cellcolor[rgb]{0.655,0.853,0.418} 55.52 & \cellcolor[rgb]{0.921,0.967,0.641} 7.32 & \cellcolor[rgb]{0.999,0.964,0.689} 30.54 & \cellcolor[rgb]{0.694,0.870,0.444} 6.46 & \cellcolor[rgb]{0.804,0.917,0.515} 30.15 & \cellcolor[rgb]{0.886,0.952,0.593} 5.92 & \cellcolor[rgb]{0.991,0.996,0.737} 40.38 & \cellcolor[rgb]{0.757,0.897,0.484} \underline{11.39} & \cellcolor[rgb]{0.968,0.986,0.705} 22.28 & \cellcolor[rgb]{0.655,0.853,0.418} 7.40 & \cellcolor[rgb]{0.725,0.883,0.464} 37.62 & \cellcolor[rgb]{0.678,0.863,0.433} 12.93 & \cellcolor[rgb]{0.827,0.927,0.530} 24.27 & \cellcolor[rgb]{0.710,0.876,0.454} 7.95 & \cellcolor[rgb]{0.827,0.927,0.530} 34.39 & \cellcolor[rgb]{0.909,0.962,0.625} 3.29 \\
MapAnything$^{*}$ & \cellcolor[rgb]{0.991,0.677,0.378} 10.19 & \cellcolor[rgb]{0.991,0.677,0.378} 14.51 & \cellcolor[rgb]{0.980,0.991,0.721} 7.92 & \cellcolor[rgb]{0.996,0.878,0.545} 23.79 & \cellcolor[rgb]{0.992,0.694,0.390} 11.03 & \cellcolor[rgb]{0.991,0.677,0.378} 12.86 & \cellcolor[rgb]{0.992,0.694,0.390} 8.86 & \cellcolor[rgb]{0.994,0.755,0.442} 27.77 & \cellcolor[rgb]{0.998,0.926,0.625} 14.81 & \cellcolor[rgb]{0.996,0.863,0.532} 16.55 & \cellcolor[rgb]{0.904,0.959,0.617} 10.64 & \cellcolor[rgb]{0.996,0.883,0.553} 22.47 & \cellcolor[rgb]{0.995,0.802,0.481} 39.29 & \cellcolor[rgb]{0.993,0.732,0.422} 2.35 & \cellcolor[rgb]{0.995,0.825,0.500} 14.68 & \cellcolor[rgb]{0.994,0.771,0.455} 17.19 & \cellcolor[rgb]{0.994,0.771,0.455} 6.00 \\
UniDepthV2$^{*}$ & \cellcolor[rgb]{0.757,0.897,0.484} 4.99 & \cellcolor[rgb]{0.857,0.940,0.553} 45.09 & \cellcolor[rgb]{0.646,0.849,0.415} \textbf{5.04} & \cellcolor[rgb]{0.646,0.849,0.415} \textbf{52.51} & \cellcolor[rgb]{0.998,0.926,0.625} 9.37 & \cellcolor[rgb]{0.997,0.921,0.617} 19.99 & \cellcolor[rgb]{0.646,0.849,0.415} \textbf{4.57} & \cellcolor[rgb]{0.646,0.849,0.415} \textbf{55.16} & \cellcolor[rgb]{0.812,0.920,0.520} 11.85 & \cellcolor[rgb]{0.646,0.849,0.415} \textbf{30.26} & \cellcolor[rgb]{0.997,0.907,0.593} 13.96 & \cellcolor[rgb]{0.796,0.914,0.510} 35.57 & \cellcolor[rgb]{0.741,0.890,0.474} 15.66 & \cellcolor[rgb]{0.996,0.888,0.561} 8.87 & \cellcolor[rgb]{0.835,0.930,0.535} 9.35 & \cellcolor[rgb]{0.796,0.914,0.510} \underline{35.35} & \cellcolor[rgb]{0.909,0.962,0.625} 3.29 \\
MoGe2$^{*}$ & \cellcolor[rgb]{0.655,0.853,0.418} \underline{4.20} & \cellcolor[rgb]{0.655,0.853,0.418} \underline{55.71} & \cellcolor[rgb]{0.944,0.977,0.673} 7.56 & \cellcolor[rgb]{0.998,0.940,0.649} 28.63 & \cellcolor[rgb]{0.678,0.863,0.433} \underline{6.36} & \cellcolor[rgb]{0.733,0.887,0.469} \underline{31.98} & \cellcolor[rgb]{0.892,0.954,0.601} 5.98 & \cellcolor[rgb]{0.997,0.999,0.745} 40.08 & \cellcolor[rgb]{0.757,0.897,0.484} 11.41 & \cellcolor[rgb]{0.944,0.977,0.673} 22.94 & \cellcolor[rgb]{0.655,0.853,0.418} \underline{7.39} & \cellcolor[rgb]{0.702,0.873,0.449} \underline{38.41} & \cellcolor[rgb]{0.671,0.859,0.428} \underline{12.66} & \cellcolor[rgb]{0.765,0.900,0.489} \underline{26.86} & \cellcolor[rgb]{0.710,0.876,0.454} \underline{7.94} & \cellcolor[rgb]{0.804,0.917,0.515} 34.95 & \cellcolor[rgb]{0.880,0.950,0.585} \underline{3.00} \\
Ours$^{*}$ & \cellcolor[rgb]{0.646,0.849,0.415} \textbf{4.16} & \cellcolor[rgb]{0.646,0.849,0.415} \textbf{55.95} & \cellcolor[rgb]{0.663,0.856,0.423} \underline{5.11} & \cellcolor[rgb]{0.686,0.866,0.439} \underline{50.64} & \cellcolor[rgb]{0.646,0.849,0.415} \textbf{6.20} & \cellcolor[rgb]{0.646,0.849,0.415} \textbf{34.16} & \cellcolor[rgb]{0.851,0.937,0.545} 5.66 & \cellcolor[rgb]{0.974,0.989,0.713} 41.32 & \cellcolor[rgb]{0.646,0.849,0.415} \textbf{10.53} & \cellcolor[rgb]{0.804,0.917,0.515} \underline{26.74} & \cellcolor[rgb]{0.646,0.849,0.415} \textbf{7.36} & \cellcolor[rgb]{0.646,0.849,0.415} \textbf{39.93} & \cellcolor[rgb]{0.646,0.849,0.415} \textbf{11.81} & \cellcolor[rgb]{0.646,0.849,0.415} \textbf{31.21} & \cellcolor[rgb]{0.646,0.849,0.415} \textbf{7.26} & \cellcolor[rgb]{0.646,0.849,0.415} \textbf{39.99} & \cellcolor[rgb]{0.646,0.849,0.415} \textbf{1.50} \\
\midrule
\multicolumn{18}{c}{\textit{Depth (metric)}} \\
\midrule
DA3 & \cellcolor[rgb]{0.710,0.876,0.454} 7.11 & \cellcolor[rgb]{0.892,0.954,0.601} 27.64 & \cellcolor[rgb]{0.702,0.873,0.449} 7.29 & \cellcolor[rgb]{0.863,0.942,0.561} 28.75 & \cellcolor[rgb]{0.869,0.945,0.569} 12.47 & \cellcolor[rgb]{0.915,0.964,0.633} 16.95 & \cellcolor[rgb]{0.646,0.849,0.415} \textbf{7.50} & \cellcolor[rgb]{0.796,0.914,0.510} \underline{31.36} & \cellcolor[rgb]{0.686,0.866,0.439} \underline{12.42} & \cellcolor[rgb]{0.646,0.849,0.415} \textbf{26.94} & \cellcolor[rgb]{0.671,0.859,0.428} 15.21 & \cellcolor[rgb]{0.646,0.849,0.415} \textbf{16.69} & \cellcolor[rgb]{0.671,0.859,0.428} 27.44 & \cellcolor[rgb]{0.996,0.863,0.532} 6.23 & \cellcolor[rgb]{0.686,0.866,0.439} \underline{12.78} & \cellcolor[rgb]{0.780,0.907,0.499} 22.08 & \cellcolor[rgb]{0.757,0.897,0.484} \underline{2.79} \\
MapAnything & \cellcolor[rgb]{0.991,0.677,0.378} 20.76 & \cellcolor[rgb]{0.992,0.686,0.384} 6.77 & \cellcolor[rgb]{0.991,0.677,0.378} 26.97 & \cellcolor[rgb]{0.991,0.677,0.378} 0.85 & \cellcolor[rgb]{0.996,0.878,0.545} 17.85 & \cellcolor[rgb]{0.991,0.677,0.378} 7.49 & \cellcolor[rgb]{0.994,0.763,0.448} 16.08 & \cellcolor[rgb]{0.991,0.677,0.378} 12.79 & \cellcolor[rgb]{0.991,0.677,0.378} 22.99 & \cellcolor[rgb]{0.993,0.717,0.409} 10.57 & \cellcolor[rgb]{0.927,0.969,0.649} 25.08 & \cellcolor[rgb]{0.993,0.702,0.397} 6.70 & \cellcolor[rgb]{0.998,0.931,0.633} 62.28 & \cellcolor[rgb]{0.994,0.755,0.442} 3.55 & \cellcolor[rgb]{0.991,0.677,0.378} 27.43 & \cellcolor[rgb]{0.991,0.677,0.378} 6.96 & \cellcolor[rgb]{0.991,0.677,0.378} 7.36 \\
UniDepthV2 & \cellcolor[rgb]{0.886,0.952,0.593} 10.59 & \cellcolor[rgb]{0.999,0.955,0.673} 19.24 & \cellcolor[rgb]{0.749,0.893,0.479} 8.58 & \cellcolor[rgb]{0.933,0.972,0.657} 24.41 & \cellcolor[rgb]{0.991,0.677,0.378} 20.71 & \cellcolor[rgb]{0.993,0.717,0.409} 8.16 & \cellcolor[rgb]{0.820,0.924,0.525} 9.53 & \cellcolor[rgb]{0.999,0.959,0.681} 22.26 & \cellcolor[rgb]{0.999,0.955,0.673} 18.45 & \cellcolor[rgb]{0.991,0.677,0.378} 9.75 & \cellcolor[rgb]{0.991,0.677,0.378} 42.98 & \cellcolor[rgb]{0.991,0.677,0.378} 6.44 & \cellcolor[rgb]{0.820,0.924,0.525} 38.15 & \cellcolor[rgb]{0.994,0.794,0.474} 4.43 & \cellcolor[rgb]{0.998,0.950,0.665} 21.28 & \cellcolor[rgb]{0.998,0.936,0.641} 13.53 & \cellcolor[rgb]{0.995,0.848,0.519} 6.21 \\
MoGe2 & \cellcolor[rgb]{0.702,0.873,0.449} 6.89 & \cellcolor[rgb]{0.804,0.917,0.515} 31.76 & \cellcolor[rgb]{0.999,0.974,0.705} 17.57 & \cellcolor[rgb]{0.993,0.732,0.422} 3.22 & \cellcolor[rgb]{0.702,0.873,0.449} 9.98 & \cellcolor[rgb]{0.646,0.849,0.415} \textbf{22.17} & \cellcolor[rgb]{0.996,0.888,0.561} 14.53 & \cellcolor[rgb]{0.993,0.709,0.403} 13.65 & \cellcolor[rgb]{0.898,0.957,0.609} 15.61 & \cellcolor[rgb]{0.998,0.945,0.657} 16.45 & \cellcolor[rgb]{0.718,0.880,0.459} 16.72 & \cellcolor[rgb]{0.985,0.994,0.729} 11.82 & \cellcolor[rgb]{0.646,0.849,0.415} \textbf{26.02} & \cellcolor[rgb]{0.995,0.817,0.493} 5.08 & \cellcolor[rgb]{0.820,0.924,0.525} 15.33 & \cellcolor[rgb]{0.999,0.969,0.697} 14.88 & \cellcolor[rgb]{0.950,0.979,0.681} 4.29 \\
MapAnything$^{*}$ & \cellcolor[rgb]{0.993,0.709,0.403} 20.26 & \cellcolor[rgb]{0.991,0.677,0.378} 6.65 & \cellcolor[rgb]{0.915,0.964,0.633} 13.60 & \cellcolor[rgb]{0.995,0.817,0.493} 7.49 & \cellcolor[rgb]{0.997,0.912,0.601} 17.10 & \cellcolor[rgb]{0.994,0.786,0.468} 9.45 & \cellcolor[rgb]{0.991,0.677,0.378} 17.11 & \cellcolor[rgb]{0.995,0.817,0.493} 16.68 & \cellcolor[rgb]{0.999,0.964,0.689} 18.25 & \cellcolor[rgb]{0.994,0.786,0.468} 12.06 & \cellcolor[rgb]{0.886,0.952,0.593} 23.34 & \cellcolor[rgb]{0.994,0.786,0.468} 7.77 & \cellcolor[rgb]{0.991,0.677,0.378} 82.54 & \cellcolor[rgb]{0.991,0.677,0.378} 1.55 & \cellcolor[rgb]{0.991,0.677,0.378} 27.46 & \cellcolor[rgb]{0.994,0.763,0.448} 8.81 & \cellcolor[rgb]{0.995,0.802,0.481} 6.57 \\
UniDepthV2$^{*}$ & \cellcolor[rgb]{0.749,0.893,0.479} 7.81 & \cellcolor[rgb]{0.880,0.950,0.585} 28.16 & \cellcolor[rgb]{0.646,0.849,0.415} \textbf{5.98} & \cellcolor[rgb]{0.646,0.849,0.415} \textbf{39.29} & \cellcolor[rgb]{1.000,0.998,0.745} 15.02 & \cellcolor[rgb]{0.997,0.907,0.593} 12.05 & \cellcolor[rgb]{0.671,0.859,0.428} \underline{7.71} & \cellcolor[rgb]{0.646,0.849,0.415} \textbf{35.58} & \cellcolor[rgb]{0.812,0.920,0.520} 14.13 & \cellcolor[rgb]{0.998,0.936,0.641} 16.17 & \cellcolor[rgb]{0.993,0.732,0.422} 40.98 & \cellcolor[rgb]{0.804,0.917,0.515} 14.75 & \cellcolor[rgb]{0.820,0.924,0.525} 37.73 & \cellcolor[rgb]{0.994,0.763,0.448} 3.64 & \cellcolor[rgb]{0.950,0.979,0.681} 18.48 & \cellcolor[rgb]{0.812,0.920,0.520} 21.38 & \cellcolor[rgb]{0.904,0.959,0.617} 3.86 \\
MoGe2$^{*}$ & \cellcolor[rgb]{0.646,0.849,0.415} \textbf{5.97} & \cellcolor[rgb]{0.646,0.849,0.415} \textbf{37.56} & \cellcolor[rgb]{0.749,0.893,0.479} 8.58 & \cellcolor[rgb]{0.909,0.962,0.625} 25.94 & \cellcolor[rgb]{0.646,0.849,0.415} \textbf{9.28} & \cellcolor[rgb]{0.749,0.893,0.479} 20.28 & \cellcolor[rgb]{0.880,0.950,0.585} 10.37 & \cellcolor[rgb]{0.950,0.979,0.681} 26.05 & \cellcolor[rgb]{0.725,0.883,0.464} 13.03 & \cellcolor[rgb]{0.927,0.969,0.649} 20.46 & \cellcolor[rgb]{0.663,0.856,0.423} \underline{14.95} & \cellcolor[rgb]{0.804,0.917,0.515} 14.74 & \cellcolor[rgb]{0.702,0.873,0.449} 29.78 & \cellcolor[rgb]{0.915,0.964,0.633} \underline{14.46} & \cellcolor[rgb]{0.702,0.873,0.449} 13.14 & \cellcolor[rgb]{0.749,0.893,0.479} \underline{22.78} & \cellcolor[rgb]{0.773,0.903,0.494} 2.86 \\
Ours$^{*}$ & \cellcolor[rgb]{0.646,0.849,0.415} \underline{5.98} & \cellcolor[rgb]{0.663,0.856,0.423} \underline{37.09} & \cellcolor[rgb]{0.671,0.859,0.428} \underline{6.47} & \cellcolor[rgb]{0.804,0.917,0.515} \underline{31.72} & \cellcolor[rgb]{0.655,0.853,0.418} \underline{9.32} & \cellcolor[rgb]{0.694,0.870,0.444} \underline{21.36} & \cellcolor[rgb]{0.874,0.947,0.577} 10.26 & \cellcolor[rgb]{0.968,0.986,0.705} 25.36 & \cellcolor[rgb]{0.646,0.849,0.415} \textbf{11.94} & \cellcolor[rgb]{0.863,0.942,0.561} \underline{22.27} & \cellcolor[rgb]{0.646,0.849,0.415} \textbf{14.38} & \cellcolor[rgb]{0.725,0.883,0.464} \underline{15.75} & \cellcolor[rgb]{0.655,0.853,0.418} \underline{26.45} & \cellcolor[rgb]{0.646,0.849,0.415} \textbf{21.58} & \cellcolor[rgb]{0.646,0.849,0.415} \textbf{12.11} & \cellcolor[rgb]{0.646,0.849,0.415} \textbf{25.02} & \cellcolor[rgb]{0.646,0.849,0.415} \textbf{2.07} \\
\bottomrule
\end{tabular}
}

\end{minipage}

\camrev{\textbf{Single-image metric accuracy.} In Table~\ref{tab:singleframe_metric}, adapting MoGe2$^{*}$ lowers average Point/Depth Rel from $7.94/13.14$ to $7.26/12.11$ ($8.6\%/7.8\%$). Higher threshold accuracy and the lowest average ranks in both blocks confirm improved absolute geometry beyond the oracle post-hoc solve.}

\par\smallskip
\noindent\begin{minipage}{\linewidth}
\centering
\scriptsize
\setlength{\tabcolsep}{1.6pt}
\renewcommand{\arraystretch}{1.0}
\captionof{table}{\camrev{Sequence metric accuracy on all seven benchmarks, including ScanNet.}}
\label{tab:sequence_metric}
\fitdetailtable{%
\begin{tabular}{l|cc|cc|cc|cc|cc|cc|cc|ccc}
\toprule
\multirow{2}{*}{Method} & \multicolumn{2}{c|}{7-Scenes} & \multicolumn{2}{c|}{ScanNet} & \multicolumn{2}{c|}{ETH3D} & \multicolumn{2}{c|}{KITTI} & \multicolumn{2}{c|}{Metropolis} & \multicolumn{2}{c|}{DIODE} & \multicolumn{2}{c|}{HAMMER} & \multicolumn{3}{c}{Avg.} \\
\cmidrule(lr){2-3}\cmidrule(lr){4-5}\cmidrule(lr){6-7}\cmidrule(lr){8-9}\cmidrule(lr){10-11}\cmidrule(lr){12-13}\cmidrule(lr){14-15}\cmidrule(lr){16-18}
 & Rel$\downarrow$ & $\delta^{1.03}\!\uparrow$ & Rel$\downarrow$ & $\delta^{1.03}\!\uparrow$ & Rel$\downarrow$ & $\delta^{1.03}\!\uparrow$ & Rel$\downarrow$ & $\delta^{1.03}\!\uparrow$ & Rel$\downarrow$ & $\delta^{1.03}\!\uparrow$ & Rel$\downarrow$ & $\delta^{1.03}\!\uparrow$ & Rel$\downarrow$ & $\delta^{1.03}\!\uparrow$ & Rel$\downarrow$ & $\delta^{1.03}\!\uparrow$ & Rk.$\downarrow$ \\
\midrule
\multicolumn{18}{c}{\textit{Point map (metric)}} \\
\midrule
UniDepthV2 & \cellcolor[rgb]{0.617,0.836,0.412} \underline{7.31} & \cellcolor[rgb]{0.663,0.856,0.423} 37.80 & \cellcolor[rgb]{0.587,0.823,0.409} 4.40 & \cellcolor[rgb]{0.607,0.832,0.411} 56.50 & \cellcolor[rgb]{0.607,0.832,0.411} 9.94 & \cellcolor[rgb]{0.997,0.921,0.617} 18.63 & \cellcolor[rgb]{0.671,0.859,0.428} 5.80 & \cellcolor[rgb]{0.765,0.900,0.489} 45.28 & \cellcolor[rgb]{0.983,0.617,0.350} 42.29 & \cellcolor[rgb]{0.983,0.617,0.350} 1.17 & \cellcolor[rgb]{0.983,0.617,0.350} 17.48 & \cellcolor[rgb]{0.983,0.617,0.350} 16.45 & \cellcolor[rgb]{0.686,0.866,0.439} 12.14 & \cellcolor[rgb]{0.997,0.902,0.585} 12.83 & \cellcolor[rgb]{0.718,0.880,0.459} 14.20 & \cellcolor[rgb]{0.933,0.972,0.657} 26.95 & \cellcolor[rgb]{0.999,0.955,0.673} 5.00 \\
MoGe2 & \cellcolor[rgb]{0.663,0.856,0.423} 8.29 & \cellcolor[rgb]{0.892,0.954,0.601} 27.05 & \cellcolor[rgb]{0.710,0.876,0.454} 6.33 & \cellcolor[rgb]{0.835,0.930,0.535} 44.24 & \cellcolor[rgb]{0.587,0.823,0.409} 5.89 & \cellcolor[rgb]{0.804,0.917,0.515} 34.29 & \cellcolor[rgb]{0.857,0.940,0.553} 7.62 & \cellcolor[rgb]{0.999,0.969,0.697} 29.61 & \cellcolor[rgb]{0.678,0.863,0.433} 17.59 & \cellcolor[rgb]{0.999,0.969,0.697} 10.84 & \cellcolor[rgb]{0.626,0.840,0.413} 7.51 & \cellcolor[rgb]{0.820,0.924,0.525} 36.25 & \cellcolor[rgb]{0.626,0.840,0.413} 9.98 & \cellcolor[rgb]{0.851,0.937,0.545} 27.94 & \cellcolor[rgb]{0.607,0.832,0.411} 9.03 & \cellcolor[rgb]{0.863,0.942,0.561} 30.03 & \cellcolor[rgb]{0.991,0.996,0.737} 4.50 \\
MapAnything & \cellcolor[rgb]{0.994,0.786,0.468} 24.51 & \cellcolor[rgb]{0.985,0.627,0.355} 1.26 & \cellcolor[rgb]{0.987,0.647,0.364} 19.73 & \cellcolor[rgb]{0.983,0.617,0.350} 6.10 & \cellcolor[rgb]{0.983,0.617,0.350} 240.27 & \cellcolor[rgb]{0.983,0.617,0.350} 5.13 & \cellcolor[rgb]{1.000,0.988,0.729} 9.59 & \cellcolor[rgb]{0.994,0.763,0.448} 17.39 & \cellcolor[rgb]{0.994,0.794,0.474} 37.22 & \cellcolor[rgb]{0.956,0.982,0.689} 13.47 & \cellcolor[rgb]{0.851,0.937,0.545} 10.08 & \cellcolor[rgb]{0.996,0.863,0.532} 23.30 & \cellcolor[rgb]{0.987,0.647,0.364} 42.10 & \cellcolor[rgb]{0.983,0.617,0.350} 0.42 & \cellcolor[rgb]{0.983,0.617,0.350} 54.79 & \cellcolor[rgb]{0.983,0.617,0.350} 9.58 & \cellcolor[rgb]{0.983,0.617,0.350} 7.07 \\
DA3-Nested & \cellcolor[rgb]{0.587,0.823,0.409} \textbf{6.78} & \cellcolor[rgb]{0.587,0.823,0.409} \textbf{40.47} & \cellcolor[rgb]{0.587,0.823,0.409} \underline{4.39} & \cellcolor[rgb]{0.587,0.823,0.409} \textbf{57.63} & \cellcolor[rgb]{0.587,0.823,0.409} 6.53 & \cellcolor[rgb]{0.587,0.823,0.409} \textbf{42.73} & \cellcolor[rgb]{0.983,0.617,0.350} 13.72 & \cellcolor[rgb]{0.983,0.617,0.350} 11.37 & \cellcolor[rgb]{1.000,0.993,0.737} 29.18 & \cellcolor[rgb]{0.993,0.702,0.397} 2.84 & \cellcolor[rgb]{0.749,0.893,0.479} 8.90 & \cellcolor[rgb]{0.587,0.823,0.409} \textbf{42.62} & \cellcolor[rgb]{0.587,0.823,0.409} \textbf{8.74} & \cellcolor[rgb]{0.915,0.964,0.633} 24.13 & \cellcolor[rgb]{0.663,0.856,0.423} 11.18 & \cellcolor[rgb]{0.812,0.920,0.520} 31.69 & \cellcolor[rgb]{0.857,0.940,0.553} \underline{3.50} \\
UniDepthV2$^{*}$ & \cellcolor[rgb]{0.617,0.836,0.412} 7.35 & \cellcolor[rgb]{0.686,0.866,0.439} 36.80 & \cellcolor[rgb]{0.587,0.823,0.409} \textbf{4.35} & \cellcolor[rgb]{0.597,0.827,0.410} \underline{57.34} & \cellcolor[rgb]{0.607,0.832,0.411} 10.77 & \cellcolor[rgb]{0.998,0.926,0.625} 18.76 & \cellcolor[rgb]{0.587,0.823,0.409} \textbf{5.13} & \cellcolor[rgb]{0.587,0.823,0.409} \textbf{52.68} & \cellcolor[rgb]{0.921,0.967,0.641} 25.64 & \cellcolor[rgb]{0.710,0.876,0.454} 20.15 & \cellcolor[rgb]{0.997,0.893,0.569} 14.35 & \cellcolor[rgb]{0.886,0.952,0.593} 33.94 & \cellcolor[rgb]{0.702,0.873,0.449} 12.72 & \cellcolor[rgb]{0.997,0.902,0.585} 12.58 & \cellcolor[rgb]{0.663,0.856,0.423} 11.47 & \cellcolor[rgb]{0.765,0.900,0.489} \underline{33.18} & \cellcolor[rgb]{0.921,0.967,0.641} 4.00 \\
MoGe2$^{*}$ & \cellcolor[rgb]{0.646,0.849,0.415} 7.98 & \cellcolor[rgb]{0.796,0.914,0.510} 31.76 & \cellcolor[rgb]{0.710,0.876,0.454} 6.32 & \cellcolor[rgb]{0.827,0.927,0.530} 44.70 & \cellcolor[rgb]{0.587,0.823,0.409} \underline{5.83} & \cellcolor[rgb]{0.780,0.907,0.499} 35.31 & \cellcolor[rgb]{0.880,0.950,0.585} 7.88 & \cellcolor[rgb]{0.998,0.940,0.649} 27.57 & \cellcolor[rgb]{0.646,0.849,0.415} \underline{16.61} & \cellcolor[rgb]{0.663,0.856,0.423} \underline{21.34} & \cellcolor[rgb]{0.626,0.840,0.413} \underline{7.49} & \cellcolor[rgb]{0.796,0.914,0.510} 36.93 & \cellcolor[rgb]{0.617,0.836,0.412} 9.72 & \cellcolor[rgb]{0.773,0.903,0.494} \underline{31.37} & \cellcolor[rgb]{0.607,0.832,0.411} \underline{8.83} & \cellcolor[rgb]{0.780,0.907,0.499} 32.71 & \cellcolor[rgb]{0.874,0.947,0.577} 3.64 \\
MapAnything$^{*}$ & \cellcolor[rgb]{0.983,0.617,0.350} 28.26 & \cellcolor[rgb]{0.983,0.617,0.350} 0.78 & \cellcolor[rgb]{0.983,0.617,0.350} 20.16 & \cellcolor[rgb]{0.983,0.617,0.350} 6.21 & \cellcolor[rgb]{0.996,0.871,0.539} 176.47 & \cellcolor[rgb]{0.987,0.647,0.364} 6.25 & \cellcolor[rgb]{0.835,0.930,0.535} 7.38 & \cellcolor[rgb]{0.999,0.959,0.681} 29.12 & \cellcolor[rgb]{0.921,0.967,0.641} 25.55 & \cellcolor[rgb]{0.694,0.870,0.444} 20.44 & \cellcolor[rgb]{0.843,0.934,0.540} 9.93 & \cellcolor[rgb]{0.997,0.893,0.569} 24.52 & \cellcolor[rgb]{0.983,0.617,0.350} 42.90 & \cellcolor[rgb]{0.987,0.647,0.364} 1.35 & \cellcolor[rgb]{0.995,0.832,0.506} 44.38 & \cellcolor[rgb]{0.993,0.725,0.416} 12.67 & \cellcolor[rgb]{0.994,0.786,0.468} 6.21 \\
Ours$^{*}$ & \cellcolor[rgb]{0.636,0.845,0.414} 7.70 & \cellcolor[rgb]{0.646,0.849,0.415} \underline{38.44} & \cellcolor[rgb]{0.702,0.873,0.449} 6.16 & \cellcolor[rgb]{0.820,0.924,0.525} 44.99 & \cellcolor[rgb]{0.587,0.823,0.409} \textbf{5.81} & \cellcolor[rgb]{0.757,0.897,0.484} \underline{36.04} & \cellcolor[rgb]{0.607,0.832,0.411} \underline{5.29} & \cellcolor[rgb]{0.655,0.853,0.418} \underline{50.05} & \cellcolor[rgb]{0.587,0.823,0.409} \textbf{15.14} & \cellcolor[rgb]{0.587,0.823,0.409} \textbf{22.80} & \cellcolor[rgb]{0.587,0.823,0.409} \textbf{7.16} & \cellcolor[rgb]{0.741,0.890,0.474} \underline{38.52} & \cellcolor[rgb]{0.597,0.827,0.410} \underline{9.13} & \cellcolor[rgb]{0.587,0.823,0.409} \textbf{38.54} & \cellcolor[rgb]{0.587,0.823,0.409} \textbf{8.06} & \cellcolor[rgb]{0.587,0.823,0.409} \textbf{38.48} & \cellcolor[rgb]{0.587,0.823,0.409} \textbf{2.07} \\
\midrule
\multicolumn{18}{c}{\textit{Depth (metric)}} \\
\midrule
DA3 & \cellcolor[rgb]{0.617,0.836,0.412} 10.09 & \cellcolor[rgb]{0.780,0.907,0.499} 23.17 & \cellcolor[rgb]{0.597,0.827,0.410} 7.23 & \cellcolor[rgb]{0.636,0.845,0.414} \underline{32.98} & \cellcolor[rgb]{0.607,0.832,0.411} 13.46 & \cellcolor[rgb]{0.843,0.934,0.540} 19.39 & \cellcolor[rgb]{0.646,0.849,0.415} 7.49 & \cellcolor[rgb]{0.812,0.920,0.520} 29.91 & \cellcolor[rgb]{0.812,0.920,0.520} 25.74 & \cellcolor[rgb]{0.998,0.950,0.665} 9.08 & \cellcolor[rgb]{0.655,0.853,0.418} 15.46 & \cellcolor[rgb]{0.587,0.823,0.409} \textbf{16.39} & \cellcolor[rgb]{0.655,0.853,0.418} 23.40 & \cellcolor[rgb]{0.996,0.883,0.553} 8.11 & \cellcolor[rgb]{0.636,0.845,0.414} 14.70 & \cellcolor[rgb]{0.820,0.924,0.525} 19.86 & \cellcolor[rgb]{0.827,0.927,0.530} \underline{3.64} \\
UniDepthV2 & \cellcolor[rgb]{0.636,0.845,0.414} 10.74 & \cellcolor[rgb]{0.869,0.945,0.569} 20.32 & \cellcolor[rgb]{0.597,0.827,0.410} 7.44 & \cellcolor[rgb]{0.765,0.900,0.489} 28.64 & \cellcolor[rgb]{0.671,0.859,0.428} 28.11 & \cellcolor[rgb]{0.994,0.763,0.448} 5.99 & \cellcolor[rgb]{0.655,0.853,0.418} 7.83 & \cellcolor[rgb]{0.835,0.930,0.535} 28.85 & \cellcolor[rgb]{0.983,0.617,0.350} 57.15 & \cellcolor[rgb]{0.983,0.617,0.350} 2.55 & \cellcolor[rgb]{0.983,0.617,0.350} 44.02 & \cellcolor[rgb]{0.983,0.617,0.350} 6.45 & \cellcolor[rgb]{0.741,0.890,0.474} 30.58 & \cellcolor[rgb]{0.994,0.786,0.468} 5.19 & \cellcolor[rgb]{0.820,0.924,0.525} 26.55 & \cellcolor[rgb]{0.999,0.974,0.705} 14.00 & \cellcolor[rgb]{0.997,0.898,0.577} 6.14 \\
MoGe2 & \cellcolor[rgb]{0.655,0.853,0.418} 11.97 & \cellcolor[rgb]{0.999,0.969,0.697} 13.17 & \cellcolor[rgb]{0.718,0.880,0.459} 11.99 & \cellcolor[rgb]{0.999,0.964,0.689} 16.38 & \cellcolor[rgb]{0.597,0.827,0.410} 11.39 & \cellcolor[rgb]{0.851,0.937,0.545} 19.01 & \cellcolor[rgb]{0.944,0.977,0.673} 17.96 & \cellcolor[rgb]{0.993,0.702,0.397} 3.34 & \cellcolor[rgb]{0.636,0.845,0.414} 17.95 & \cellcolor[rgb]{0.702,0.873,0.449} 16.86 & \cellcolor[rgb]{0.678,0.863,0.433} 16.31 & \cellcolor[rgb]{0.927,0.969,0.649} 12.53 & \cellcolor[rgb]{0.646,0.849,0.415} 22.94 & \cellcolor[rgb]{0.994,0.755,0.442} 4.32 & \cellcolor[rgb]{0.655,0.853,0.418} 15.79 & \cellcolor[rgb]{0.997,0.921,0.617} 12.23 & \cellcolor[rgb]{0.998,0.940,0.649} 5.71 \\
MapAnything & \cellcolor[rgb]{0.998,0.931,0.633} 42.86 & \cellcolor[rgb]{0.993,0.732,0.422} 3.64 & \cellcolor[rgb]{0.983,0.617,0.350} 45.60 & \cellcolor[rgb]{0.983,0.617,0.350} 2.54 & \cellcolor[rgb]{0.983,0.617,0.350} 263.24 & \cellcolor[rgb]{0.991,0.677,0.378} 3.82 & \cellcolor[rgb]{1.000,0.988,0.729} 21.03 & \cellcolor[rgb]{0.990,0.667,0.373} 2.09 & \cellcolor[rgb]{0.999,0.955,0.673} 40.04 & \cellcolor[rgb]{0.904,0.959,0.617} 12.98 & \cellcolor[rgb]{0.725,0.883,0.464} 17.82 & \cellcolor[rgb]{0.985,0.994,0.729} 11.60 & \cellcolor[rgb]{0.998,0.931,0.633} 65.79 & \cellcolor[rgb]{0.983,0.617,0.350} 0.67 & \cellcolor[rgb]{0.983,0.617,0.350} 70.91 & \cellcolor[rgb]{0.983,0.617,0.350} 5.33 & \cellcolor[rgb]{0.983,0.617,0.350} 7.93 \\
DA3-Nested & \cellcolor[rgb]{0.587,0.823,0.409} \textbf{8.41} & \cellcolor[rgb]{0.587,0.823,0.409} \textbf{29.00} & \cellcolor[rgb]{0.587,0.823,0.409} \textbf{6.99} & \cellcolor[rgb]{0.587,0.823,0.409} \textbf{34.49} & \cellcolor[rgb]{0.597,0.827,0.410} 10.81 & \cellcolor[rgb]{0.671,0.859,0.428} 23.63 & \cellcolor[rgb]{0.983,0.617,0.350} 34.63 & \cellcolor[rgb]{0.983,0.617,0.350} 0.26 & \cellcolor[rgb]{0.999,0.959,0.681} 39.64 & \cellcolor[rgb]{0.986,0.637,0.360} 2.77 & \cellcolor[rgb]{0.617,0.836,0.412} \underline{14.57} & \cellcolor[rgb]{0.597,0.827,0.410} \underline{16.26} & \cellcolor[rgb]{0.626,0.840,0.413} 21.11 & \cellcolor[rgb]{0.994,0.755,0.442} 4.36 & \cellcolor[rgb]{0.710,0.876,0.454} 19.45 & \cellcolor[rgb]{0.968,0.986,0.705} 15.82 & \cellcolor[rgb]{0.874,0.947,0.577} 4.00 \\
UniDepthV2$^{*}$ & \cellcolor[rgb]{0.617,0.836,0.412} \underline{9.84} & \cellcolor[rgb]{0.780,0.907,0.499} 23.22 & \cellcolor[rgb]{0.587,0.823,0.409} \underline{7.04} & \cellcolor[rgb]{0.718,0.880,0.459} 30.44 & \cellcolor[rgb]{0.626,0.840,0.413} 18.34 & \cellcolor[rgb]{0.998,0.950,0.665} 11.84 & \cellcolor[rgb]{0.587,0.823,0.409} \textbf{6.10} & \cellcolor[rgb]{0.587,0.823,0.409} \textbf{39.04} & \cellcolor[rgb]{0.820,0.924,0.525} 26.15 & \cellcolor[rgb]{0.607,0.832,0.411} \underline{18.39} & \cellcolor[rgb]{0.992,0.686,0.384} 42.21 & \cellcolor[rgb]{0.886,0.952,0.593} 13.07 & \cellcolor[rgb]{0.733,0.887,0.469} 30.17 & \cellcolor[rgb]{0.994,0.771,0.455} 4.71 & \cellcolor[rgb]{0.725,0.883,0.464} 19.98 & \cellcolor[rgb]{0.812,0.920,0.520} 20.10 & \cellcolor[rgb]{0.863,0.942,0.561} 3.93 \\
MoGe2$^{*}$ & \cellcolor[rgb]{0.663,0.856,0.423} 12.13 & \cellcolor[rgb]{0.915,0.964,0.633} 18.37 & \cellcolor[rgb]{0.655,0.853,0.418} 9.28 & \cellcolor[rgb]{0.780,0.907,0.499} 27.96 & \cellcolor[rgb]{0.587,0.823,0.409} \underline{9.15} & \cellcolor[rgb]{0.617,0.836,0.412} \underline{24.93} & \cellcolor[rgb]{0.686,0.866,0.439} 8.83 & \cellcolor[rgb]{0.880,0.950,0.585} 26.44 & \cellcolor[rgb]{0.626,0.840,0.413} \underline{17.59} & \cellcolor[rgb]{0.646,0.849,0.415} 17.85 & \cellcolor[rgb]{0.626,0.840,0.413} 14.75 & \cellcolor[rgb]{0.843,0.934,0.540} 13.69 & \cellcolor[rgb]{0.626,0.840,0.413} \underline{21.10} & \cellcolor[rgb]{0.933,0.972,0.657} \underline{16.14} & \cellcolor[rgb]{0.607,0.832,0.411} \underline{13.26} & \cellcolor[rgb]{0.780,0.907,0.499} \underline{20.77} & \cellcolor[rgb]{0.863,0.942,0.561} 3.93 \\
MapAnything$^{*}$ & \cellcolor[rgb]{0.983,0.617,0.350} 63.61 & \cellcolor[rgb]{0.983,0.617,0.350} 0.40 & \cellcolor[rgb]{0.992,0.694,0.390} 42.85 & \cellcolor[rgb]{0.985,0.627,0.355} 2.82 & \cellcolor[rgb]{0.996,0.883,0.553} 189.74 & \cellcolor[rgb]{0.983,0.617,0.350} 2.51 & \cellcolor[rgb]{0.796,0.914,0.510} 12.37 & \cellcolor[rgb]{0.997,0.893,0.569} 12.10 & \cellcolor[rgb]{0.835,0.930,0.535} 26.69 & \cellcolor[rgb]{0.694,0.870,0.444} 16.98 & \cellcolor[rgb]{0.804,0.917,0.515} 20.54 & \cellcolor[rgb]{0.999,0.979,0.713} 11.04 & \cellcolor[rgb]{0.983,0.617,0.350} 94.31 & \cellcolor[rgb]{0.983,0.617,0.350} 0.67 & \cellcolor[rgb]{0.993,0.732,0.422} 64.30 & \cellcolor[rgb]{0.992,0.694,0.390} 6.65 & \cellcolor[rgb]{0.993,0.702,0.397} 7.50 \\
Ours$^{*}$ & \cellcolor[rgb]{0.636,0.845,0.414} 10.83 & \cellcolor[rgb]{0.757,0.897,0.484} \underline{24.06} & \cellcolor[rgb]{0.646,0.849,0.415} 9.11 & \cellcolor[rgb]{0.780,0.907,0.499} 28.09 & \cellcolor[rgb]{0.587,0.823,0.409} \textbf{9.07} & \cellcolor[rgb]{0.587,0.823,0.409} \textbf{25.50} & \cellcolor[rgb]{0.597,0.827,0.410} \underline{6.47} & \cellcolor[rgb]{0.710,0.876,0.454} \underline{34.32} & \cellcolor[rgb]{0.587,0.823,0.409} \textbf{16.02} & \cellcolor[rgb]{0.587,0.823,0.409} \textbf{18.67} & \cellcolor[rgb]{0.587,0.823,0.409} \textbf{13.68} & \cellcolor[rgb]{0.718,0.880,0.459} 15.07 & \cellcolor[rgb]{0.587,0.823,0.409} \textbf{18.72} & \cellcolor[rgb]{0.587,0.823,0.409} \textbf{26.45} & \cellcolor[rgb]{0.587,0.823,0.409} \textbf{11.99} & \cellcolor[rgb]{0.587,0.823,0.409} \textbf{24.60} & \cellcolor[rgb]{0.587,0.823,0.409} \textbf{2.21} \\
\bottomrule
\end{tabular}
}

\end{minipage}

\camrev{\textbf{Sequence metric accuracy.} On all seven benchmarks, including ScanNet, Table~\ref{tab:sequence_metric} reduces MoGe2$^{*}$ Point/Depth Rel from $8.83/13.26$ to $8.06/11.99$ ($8.7\%/9.6\%$). Higher threshold accuracy and the lowest average ranks are obtained on held-out sequences without further updates, supporting transfer beyond the adaptation frames.}

\clearpage
\camrev{Tables~\ref{tab:singleframe_perdataset_delta103} and~\ref{tab:sequence_perdataset_delta103} evaluate shape after removing global scale or affine ambiguity. The single-image comparison retains all ten benchmarks; the sequence comparison uses the same seven benchmarks and held-out sequences as Table~\ref{tab:sequence_metric}. DA3 and DA3-Nested follow Depth Anything 3~\cite{lin2025da3}.}

\par\smallskip
\noindent\begin{minipage}{\linewidth}
\centering
\scriptsize
\setlength{\tabcolsep}{1.6pt}
\renewcommand{\arraystretch}{1.05}
\captionof{table}{\camrev{Single-image relative geometry on ten datasets under scale- and affine-invariant evaluation.}}
\label{tab:singleframe_perdataset_delta103}
\fitdetailtable{%
\begin{tabular}{l|cc|cc|cc|cc|cc|cc|cc|cc|cc|cc|ccc}
\toprule
\multirow{2}{*}{Method} & \multicolumn{2}{c|}{NYUv2} & \multicolumn{2}{c|}{KITTI} & \multicolumn{2}{c|}{ETH3D} & \multicolumn{2}{c|}{iBims-1} & \multicolumn{2}{c|}{GSO} & \multicolumn{2}{c|}{Sintel} & \multicolumn{2}{c|}{DDAD} & \multicolumn{2}{c|}{DIODE} & \multicolumn{2}{c|}{Spring} & \multicolumn{2}{c|}{HAMMER} & \multicolumn{3}{c}{Avg.} \\
\cmidrule(lr){2-3}\cmidrule(lr){4-5}\cmidrule(lr){6-7}\cmidrule(lr){8-9}\cmidrule(lr){10-11}\cmidrule(lr){12-13}\cmidrule(lr){14-15}\cmidrule(lr){16-17}\cmidrule(lr){18-19}\cmidrule(lr){20-21}\cmidrule(lr){22-24}
 & Rel$\downarrow$ & $\delta^{1.03}\!\uparrow$ & Rel$\downarrow$ & $\delta^{1.03}\!\uparrow$ & Rel$\downarrow$ & $\delta^{1.03}\!\uparrow$ & Rel$\downarrow$ & $\delta^{1.03}\!\uparrow$ & Rel$\downarrow$ & $\delta^{1.03}\!\uparrow$ & Rel$\downarrow$ & $\delta^{1.03}\!\uparrow$ & Rel$\downarrow$ & $\delta^{1.03}\!\uparrow$ & Rel$\downarrow$ & $\delta^{1.03}\!\uparrow$ & Rel$\downarrow$ & $\delta^{1.03}\!\uparrow$ & Rel$\downarrow$ & $\delta^{1.03}\!\uparrow$ & Rel$\downarrow$ & $\delta^{1.03}\!\uparrow$ & \rank$\downarrow$ \\
\midrule
\multicolumn{24}{c}{\textit{Point map (scale-invariant)}} \\
\midrule
DA3 & - & - & - & - & - & - & - & - & - & - & - & - & - & - & - & - & - & - & - & - & - & - & - \\
\bottomrule
\end{tabular}
}

\end{minipage}

\camrev{\textbf{Single-image relative geometry.} Table~\ref{tab:singleframe_perdataset_delta103} shows lower scale- and affine-invariant Point Rel than MoGe2$^{*}$ ($9.10\!\rightarrow\!8.56$ and $8.02\!\rightarrow\!7.73$), accompanied by higher threshold accuracy. Scale-invariant Depth Rel also falls from $9.08$ to $8.55$. Affine-invariant depth is an exception: Rel changes from $5.58$ to $5.87$ and threshold accuracy from $64.52$ to $64.13$. Overall, relative geometry performance remains comparable to the pre-adaptation baseline.}

\camrev{\textbf{Sequence relative geometry.} Table~\ref{tab:sequence_perdataset_delta103} shows that the improvement persists after removing global scale or affine ambiguity. Relative to MoGe2$^{*}$, scale- and affine-invariant Point Rel fall from $7.03/6.54$ to $6.24/5.94$, while Depth Rel falls from $7.01/5.38$ to $6.23/5.20$. Average threshold accuracy also increases in all four blocks. These aligned evaluations indicate that sequence adaptation improves the predicted shape as well as its absolute calibration, although the amount of improvement varies across datasets and alignment settings.}

\clearpage
\par\smallskip
\noindent\begin{minipage}{\linewidth}
\centering
\scriptsize
\setlength{\tabcolsep}{1.6pt}
\renewcommand{\arraystretch}{1.0}
\captionof{table}{\camrev{Sequence relative geometry under scale- and affine-invariant evaluation on seven benchmarks.}}
\label{tab:sequence_perdataset_delta103}
\fitdetailtable{%
\begin{tabular}{l|cc|cc|cc|cc|cc|cc|cc|ccc}
\toprule
\multirow{2}{*}{Method} & \multicolumn{2}{c|}{7-Scenes} & \multicolumn{2}{c|}{ScanNet} & \multicolumn{2}{c|}{ETH3D} & \multicolumn{2}{c|}{KITTI} & \multicolumn{2}{c|}{Metropolis} & \multicolumn{2}{c|}{DIODE} & \multicolumn{2}{c|}{HAMMER} & \multicolumn{3}{c}{Avg.} \\
\cmidrule(lr){2-3}\cmidrule(lr){4-5}\cmidrule(lr){6-7}\cmidrule(lr){8-9}\cmidrule(lr){10-11}\cmidrule(lr){12-13}\cmidrule(lr){14-15}\cmidrule(lr){16-18}
 & Rel$\downarrow$ & $\delta^{1.03}\!\uparrow$ & Rel$\downarrow$ & $\delta^{1.03}\!\uparrow$ & Rel$\downarrow$ & $\delta^{1.03}\!\uparrow$ & Rel$\downarrow$ & $\delta^{1.03}\!\uparrow$ & Rel$\downarrow$ & $\delta^{1.03}\!\uparrow$ & Rel$\downarrow$ & $\delta^{1.03}\!\uparrow$ & Rel$\downarrow$ & $\delta^{1.03}\!\uparrow$ & Rel$\downarrow$ & $\delta^{1.03}\!\uparrow$ & Rk.$\downarrow$ \\
\midrule
\multicolumn{18}{c}{\textit{Point map (scale-invariant)}} \\
\midrule
UniDepthV2 & \cellcolor[rgb]{0.765,0.900,0.489} 7.00 & \cellcolor[rgb]{0.863,0.942,0.561} 39.99 & \cellcolor[rgb]{0.718,0.880,0.459} 4.14 & \cellcolor[rgb]{0.757,0.897,0.484} 60.60 & \cellcolor[rgb]{0.851,0.937,0.545} 8.02 & \cellcolor[rgb]{0.996,0.855,0.526} 21.58 & \cellcolor[rgb]{0.617,0.836,0.412} 5.41 & \cellcolor[rgb]{0.886,0.952,0.593} 40.66 & \cellcolor[rgb]{0.983,0.617,0.350} 37.77 & \cellcolor[rgb]{0.983,0.617,0.350} 0.31 & \cellcolor[rgb]{0.996,0.888,0.561} 12.29 & \cellcolor[rgb]{0.983,0.617,0.350} 7.87 & \cellcolor[rgb]{0.686,0.866,0.439} 4.04 & \cellcolor[rgb]{0.804,0.917,0.515} 53.37 & \cellcolor[rgb]{0.968,0.986,0.705} 11.24 & \cellcolor[rgb]{0.998,0.940,0.649} 32.05 & \cellcolor[rgb]{0.995,0.840,0.513} 6.00 \\
\bottomrule
\end{tabular}
}

\end{minipage}

\clearpage
\subsection{\octrev{Detailed results across base models}}
\label{sec:app_multi_baseline}
\camrev{Table~\ref{tab:app_multi_baseline_perdataset} pairs MapAnything~\cite{keetha2025mapanything}, UniDepthV2~\cite{piccinelli2025unidepthv2}, and MoGe2~\cite{wang2026moge2} with their adapted variants. All use known intrinsics; Point scores precede reprojection. Bold marks the better value within each pair, including both values for ties. Global rankings are omitted.}

\camrev{Adaptation reduces average metric Point/Depth Rel from $44.30/64.21$ to $40.18/59.82$ for MapAnything, $11.48/19.98$ to $10.75/19.31$ for UniDepthV2, and $9.01/13.26$ to $8.15/11.99$ for MoGe2. Dataset-level changes are mixed for the first two models; MoGe2 improves metric Depth Rel on all seven datasets.}

\par\smallskip
\noindent\begin{minipage}{\linewidth}
\centering
\scriptsize
\setlength{\tabcolsep}{3pt}
\renewcommand{\arraystretch}{1.0}
\captionof{table}{\camrev{Paired sequence comparisons of each base model and its adapted variant; bold marks the better value within each pair.}}
\label{tab:app_multi_baseline_perdataset}
\fitdetailtable{%
\begin{tabular}{l|cc|cc|cc|cc|cc|cc|cc|cc}
\toprule
\multirow{2}{*}{Method} & \multicolumn{2}{c|}{7-Scenes} & \multicolumn{2}{c|}{ScanNet} & \multicolumn{2}{c|}{ETH3D} & \multicolumn{2}{c|}{KITTI} & \multicolumn{2}{c|}{Metropolis} & \multicolumn{2}{c|}{DIODE} & \multicolumn{2}{c|}{HAMMER} & \multicolumn{2}{c}{Avg.} \\
\cmidrule(lr){2-3}\cmidrule(lr){4-5}\cmidrule(lr){6-7}\cmidrule(lr){8-9}\cmidrule(lr){10-11}\cmidrule(lr){12-13}\cmidrule(lr){14-15}\cmidrule(lr){16-17}
 & Rel$\downarrow$ & $\delta^{1.03}\!\uparrow$ & Rel$\downarrow$ & $\delta^{1.03}\!\uparrow$ & Rel$\downarrow$ & $\delta^{1.03}\!\uparrow$ & Rel$\downarrow$ & $\delta^{1.03}\!\uparrow$ & Rel$\downarrow$ & $\delta^{1.03}\!\uparrow$ & Rel$\downarrow$ & $\delta^{1.03}\!\uparrow$ & Rel$\downarrow$ & $\delta^{1.03}\!\uparrow$ & Rel$\downarrow$ & $\delta^{1.03}\!\uparrow$ \\
\midrule
\multicolumn{17}{c}{\textit{Point map (scale-invariant)}} \\
\midrule
MapAnything$^{*}$ & 6.78 & 34.29 & 4.96 & 38.04 & \textbf{15.02} & 48.36 & \textbf{5.42} & 48.72 & 16.25 & 29.03 & 6.42 & 62.40 & 5.50 & 21.65 & 8.62 & 40.36 \\
MapAnything$^{*}$ (Ours) & \textbf{6.57} & \textbf{38.90} & \textbf{4.83} & \textbf{42.01} & 15.70 & \textbf{48.48} & 6.54 & \textbf{51.30} & \textbf{14.38} & \textbf{29.94} & \textbf{6.36} & \textbf{62.75} & \textbf{5.09} & \textbf{30.17} & \textbf{8.50} & \textbf{43.36} \\
\midrule
UniDepthV2$^{*}$ & \textbf{6.27} & \textbf{54.21} & 3.74 & 69.00 & \textbf{3.60} & 66.66 & 4.45 & 59.51 & 27.65 & 16.87 & \textbf{6.08} & \textbf{66.67} & 3.51 & \textbf{63.04} & 7.90 & \textbf{56.57} \\
UniDepthV2$^{*}$ (Ours) & 6.32 & 53.26 & \textbf{3.71} & \textbf{69.79} & 3.64 & \textbf{67.09} & \textbf{4.41} & \textbf{59.88} & \textbf{23.87} & \textbf{18.04} & 6.70 & 62.54 & \textbf{3.50} & 62.68 & \textbf{7.45} & 56.18 \\
\midrule
MoGe2$^{*}$ & 6.75 & 47.78 & 5.37 & 59.53 & \textbf{3.79} & \textbf{62.46} & 7.94 & 27.83 & 17.27 & 9.21 & 6.34 & 59.96 & 4.31 & 50.90 & 7.39 & 45.38 \\
MoGe2$^{*}$ (Ours) & \textbf{6.48} & \textbf{52.79} & \textbf{5.16} & \textbf{61.04} & 3.81 & 62.26 & \textbf{4.77} & \textbf{55.77} & \textbf{14.90} & \textbf{20.33} & \textbf{5.75} & \textbf{65.11} & \textbf{4.00} & \textbf{55.16} & \textbf{6.41} & \textbf{53.21} \\
\midrule
\multicolumn{17}{c}{\textit{Point map (affine-invariant)}} \\
\midrule
MapAnything$^{*}$ & 5.94 & \textbf{52.36} & 3.60 & 66.00 & \textbf{11.03} & \textbf{47.00} & \textbf{5.05} & 52.01 & 15.55 & 29.33 & 5.80 & 63.02 & \textbf{3.39} & \textbf{60.44} & 7.20 & 52.88 \\
MapAnything$^{*}$ (Ours) & \textbf{5.91} & 51.95 & \textbf{3.54} & \textbf{67.15} & 11.28 & 46.76 & 5.57 & \textbf{53.25} & \textbf{14.06} & \textbf{30.13} & \textbf{5.70} & \textbf{63.23} & 3.40 & 60.17 & \textbf{7.07} & \textbf{53.23} \\
\midrule
UniDepthV2$^{*}$ & \textbf{6.10} & \textbf{56.25} & \textbf{3.31} & 72.21 & \textbf{3.19} & \textbf{68.92} & 4.26 & 62.29 & 27.00 & 13.85 & \textbf{5.43} & \textbf{67.10} & 3.07 & \textbf{67.50} & 7.48 & \textbf{58.30} \\
UniDepthV2$^{*}$ (Ours) & 6.15 & 55.18 & \textbf{3.31} & \textbf{72.67} & 3.22 & 68.88 & \textbf{4.25} & \textbf{62.35} & \textbf{23.77} & \textbf{16.29} & 6.01 & 62.21 & \textbf{3.06} & 67.35 & \textbf{7.11} & 57.85 \\
\midrule
MoGe2$^{*}$ & 6.63 & 49.27 & 5.13 & 60.78 & \textbf{3.61} & \textbf{62.39} & 6.60 & 31.82 & 16.70 & 15.73 & 5.30 & 60.29 & 4.17 & 53.00 & 6.88 & 47.61 \\
MoGe2$^{*}$ (Ours) & \textbf{6.41} & \textbf{53.43} & \textbf{4.93} & \textbf{61.90} & 3.62 & 62.11 & \textbf{4.42} & \textbf{58.39} & \textbf{14.47} & \textbf{24.52} & \textbf{4.81} & \textbf{64.83} & \textbf{3.86} & \textbf{56.82} & \textbf{6.07} & \textbf{54.57} \\
\midrule
\multicolumn{17}{c}{\textit{Point map (metric)}} \\
\midrule
MapAnything$^{*}$ & \textbf{28.20} & \textbf{0.78} & 20.16 & 6.17 & 176.03 & \textbf{6.26} & \textbf{7.36} & 29.02 & 25.55 & 20.44 & 9.93 & 24.49 & 42.89 & 1.35 & 44.30 & 12.64 \\
MapAnything$^{*}$ (Ours) & 30.49 & 0.56 & \textbf{17.53} & \textbf{6.90} & \textbf{157.33} & 5.04 & 8.53 & \textbf{35.46} & \textbf{15.84} & \textbf{21.84} & \textbf{9.78} & \textbf{24.52} & \textbf{41.73} & \textbf{2.72} & \textbf{40.18} & \textbf{13.86} \\
\midrule
UniDepthV2$^{*}$ & 7.35 & 36.80 & \textbf{4.35} & \textbf{57.34} & 10.77 & 18.76 & \textbf{5.13} & \textbf{52.67} & 25.64 & \textbf{20.15} & \textbf{14.35} & \textbf{33.94} & 12.74 & 12.57 & 11.48 & \textbf{33.18} \\
UniDepthV2$^{*}$ (Ours) & \textbf{7.16} & \textbf{38.96} & 4.38 & 56.92 & \textbf{9.83} & \textbf{19.69} & 5.16 & 51.94 & \textbf{22.48} & 15.94 & 15.11 & 26.87 & \textbf{11.14} & \textbf{20.08} & \textbf{10.75} & 32.92 \\
\midrule
MoGe2$^{*}$ & 8.29 & 27.03 & 6.33 & 44.20 & \textbf{5.90} & 33.96 & 7.61 & 29.62 & 17.47 & 11.37 & 7.50 & 36.30 & 9.97 & 28.00 & 9.01 & 30.07 \\
MoGe2$^{*}$ (Ours) & \textbf{7.79} & \textbf{36.66} & \textbf{6.16} & \textbf{44.68} & \textbf{5.90} & \textbf{34.14} & \textbf{5.29} & \textbf{49.96} & \textbf{15.51} & \textbf{17.52} & \textbf{7.20} & \textbf{38.10} & \textbf{9.20} & \textbf{36.67} & \textbf{8.15} & \textbf{36.82} \\
\midrule
\multicolumn{17}{c}{\textit{Depth (scale-invariant)}} \\
\midrule
MapAnything$^{*}$ & 5.85 & \textbf{53.52} & 3.74 & 64.42 & \textbf{14.89} & \textbf{49.39} & \textbf{5.21} & \textbf{52.08} & 16.14 & 29.97 & 6.29 & 63.22 & \textbf{3.38} & \textbf{60.97} & \textbf{7.93} & 53.37 \\
MapAnything$^{*}$ (Ours) & \textbf{5.84} & 52.90 & \textbf{3.68} & \textbf{65.55} & 15.52 & \textbf{49.39} & 6.49 & 51.94 & \textbf{14.29} & \textbf{30.88} & \textbf{6.25} & \textbf{63.49} & 3.40 & 60.44 & \textbf{7.93} & \textbf{53.51} \\
\midrule
UniDepthV2$^{*}$ & \textbf{6.13} & \textbf{55.48} & 3.54 & 70.16 & \textbf{3.14} & 69.76 & 4.41 & 59.67 & 26.84 & 20.33 & \textbf{6.08} & \textbf{66.66} & \textbf{3.15} & \textbf{66.19} & 7.61 & \textbf{58.32} \\
UniDepthV2$^{*}$ (Ours) & 6.17 & 54.56 & \textbf{3.51} & \textbf{71.02} & 3.16 & \textbf{70.30} & \textbf{4.37} & \textbf{60.04} & \textbf{23.06} & \textbf{21.80} & 6.70 & 62.45 & 3.16 & 65.98 & \textbf{7.16} & 58.02 \\
\midrule
MoGe2$^{*}$ & 6.49 & 51.08 & 5.27 & 61.58 & \textbf{3.66} & \textbf{63.29} & 8.08 & 29.78 & 15.32 & 31.20 & 6.13 & 63.98 & 4.11 & 55.84 & 7.01 & 50.96 \\
MoGe2$^{*}$ (Ours) & \textbf{6.41} & \textbf{53.12} & \textbf{5.09} & \textbf{62.47} & 3.67 & 63.24 & \textbf{4.76} & \textbf{55.93} & \textbf{14.10} & \textbf{31.80} & \textbf{5.63} & \textbf{66.25} & \textbf{3.96} & \textbf{56.64} & \textbf{6.23} & \textbf{55.63} \\
\midrule
\multicolumn{17}{c}{\textit{Depth (affine-invariant)}} \\
\midrule
MapAnything$^{*}$ & 5.62 & \textbf{57.84} & 2.93 & 73.89 & \textbf{7.84} & \textbf{55.35} & \textbf{4.78} & 54.68 & 15.04 & 32.40 & 4.96 & 71.25 & \textbf{2.62} & \textbf{73.03} & 6.26 & 59.78 \\
MapAnything$^{*}$ (Ours) & \textbf{5.61} & 57.32 & \textbf{2.88} & \textbf{75.07} & 8.02 & 55.25 & 5.20 & \textbf{55.00} & \textbf{13.67} & \textbf{33.39} & \textbf{4.80} & \textbf{71.33} & \textbf{2.62} & 72.41 & \textbf{6.12} & \textbf{59.97} \\
\midrule
UniDepthV2$^{*}$ & \textbf{5.82} & \textbf{59.25} & \textbf{2.66} & \textbf{79.76} & 2.68 & 76.13 & \textbf{4.08} & \textbf{64.22} & 26.75 & 17.35 & \textbf{4.31} & \textbf{74.63} & 2.32 & 76.19 & 6.94 & 63.93 \\
UniDepthV2$^{*}$ (Ours) & 5.86 & 58.76 & 2.67 & 79.60 & \textbf{2.67} & \textbf{76.56} & \textbf{4.08} & 64.06 & \textbf{23.59} & \textbf{19.53} & 4.52 & 72.81 & \textbf{2.31} & \textbf{76.24} & \textbf{6.53} & \textbf{63.94} \\
\midrule
MoGe2$^{*}$ & \textbf{5.96} & 59.12 & 4.26 & 68.34 & \textbf{2.66} & \textbf{74.32} & 3.85 & 63.85 & 14.72 & 33.71 & \textbf{3.40} & \textbf{76.00} & \textbf{2.78} & \textbf{71.99} & 5.38 & 63.90 \\
MoGe2$^{*}$ (Ours) & 5.97 & \textbf{59.13} & \textbf{4.11} & \textbf{69.11} & 2.67 & 74.15 & \textbf{3.83} & \textbf{64.12} & \textbf{13.58} & \textbf{34.25} & 3.42 & 75.54 & 2.79 & 71.57 & \textbf{5.20} & \textbf{63.98} \\
\midrule
\multicolumn{17}{c}{\textit{Depth (metric)}} \\
\midrule
MapAnything$^{*}$ & \textbf{63.43} & \textbf{0.40} & 42.84 & 2.79 & 189.30 & \textbf{2.49} & 12.36 & 11.99 & 26.69 & 16.98 & 20.55 & 11.02 & 94.30 & 0.71 & 64.21 & 6.63 \\
MapAnything$^{*}$ (Ours) & 69.06 & 0.38 & \textbf{37.07} & \textbf{2.98} & \textbf{171.01} & 2.19 & \textbf{11.73} & \textbf{20.78} & \textbf{16.73} & \textbf{18.48} & \textbf{20.38} & \textbf{11.49} & \textbf{92.74} & \textbf{1.78} & \textbf{59.82} & \textbf{8.30} \\
\midrule
UniDepthV2$^{*}$ & 9.84 & 23.22 & \textbf{7.04} & \textbf{30.44} & 18.33 & \textbf{11.87} & \textbf{6.10} & \textbf{39.03} & 26.15 & \textbf{18.39} & \textbf{42.21} & \textbf{13.06} & 30.21 & 4.72 & 19.98 & \textbf{20.10} \\
UniDepthV2$^{*}$ (Ours) & \textbf{9.36} & \textbf{26.78} & 7.17 & 29.08 & \textbf{16.72} & 11.67 & 6.26 & 36.82 & \textbf{23.77} & 13.72 & 46.10 & 9.84 & \textbf{25.77} & \textbf{11.75} & \textbf{19.31} & 19.95 \\
\midrule
MoGe2$^{*}$ & 12.13 & 18.37 & 9.28 & 27.96 & 9.15 & 24.93 & 8.83 & 26.44 & 17.59 & 17.85 & 14.75 & 13.69 & 21.10 & 16.14 & 13.26 & 20.77 \\
MoGe2$^{*}$ (Ours) & \textbf{10.83} & \textbf{24.06} & \textbf{9.11} & \textbf{28.09} & \textbf{9.07} & \textbf{25.50} & \textbf{6.47} & \textbf{34.32} & \textbf{16.02} & \textbf{18.67} & \textbf{13.68} & \textbf{15.07} & \textbf{18.72} & \textbf{26.45} & \textbf{11.99} & \textbf{24.60} \\
\bottomrule
\end{tabular}
}

\end{minipage}

\end{document}